\documentclass[conference]{IEEEtran}
\IEEEoverridecommandlockouts

\usepackage[
style=authoryear,
backend=biber,
maxcitenames=2,
style=numeric,
citestyle=numeric,
sorting=none,
doi=false,
isbn=false,
url=false,
eprint=true]{biblatex}
\usepackage{amsmath,amssymb,amsfonts}
\usepackage{algorithmic}
\usepackage{graphicx}
\usepackage{textcomp}
\usepackage{xcolor}
\usepackage{subcaption}

\usepackage{comment}

\def\BibTeX{{\rm B\kern-.05em{\sc i\kern-.025em b}\kern-.08em
   T\kern-.1667em\lower.7ex\hbox{E}\kern-.125emX}}
    
\begin{document}

\title{Fault injection analysis of Real NVP normalising flow model for satellite anomaly detection\\
\thanks{This publication is part of the project PNRR-NGEU which has received funding from the MUR – DM 117/2023 and MUR – DM 351/2022.

$^\dag$ Corresponding author}
}

\author{\IEEEauthorblockN{Gabriele Greco}
\IEEEauthorblockA{
\textit{DET - Politecnico di Torino}\\
Torino, ITA \\
gabriele.greco@studenti.polito.it}
\and
\IEEEauthorblockN{Carlo Cena $^\dag$}
\IEEEauthorblockA{
\textit{DET - Politecnico di Torino}\\
Torino, ITA \\
carlo.cena@polito.it}
\and
\IEEEauthorblockN{Umberto Albertin}
\IEEEauthorblockA{
\textit{DET - Politecnico di Torino}\\
Torino, ITA \\
umberto.albertin@polito.it}
\newlineauthors
\IEEEauthorblockN{Mauro Martini}
\IEEEauthorblockA{
\textit{DET - Politecnico di Torino}\\
Torino, ITA \\
mauro.martini@polito.it}
\and
\IEEEauthorblockN{Marcello Chiaberge}
\IEEEauthorblockA{
\textit{DET - Politecnico di Torino}\\
Torino, ITA \\
marcello.chiaberge@polito.it}
}

\maketitle

\begin{abstract}
Satellites are used for a multitude of applications, including communications, Earth observation, and space science. Neural networks and deep learning-based approaches now represent the state-of-the-art to enhance the performance and efficiency of these tasks. Given that satellites are susceptible to various faults, one critical application of Artificial Intelligence (AI) is fault detection. However, despite the advantages of neural networks, these systems are vulnerable to radiation errors, which can significantly impact their reliability. Ensuring the dependability of these solutions requires extensive testing and validation, particularly using fault injection methods. This study analyses a physics-informed (PI) real-valued non-volume preserving (Real NVP) normalizing flow model for fault detection in space systems, with a focus on resilience to Single-Event Upsets (SEUs). We present a customized fault injection framework in TensorFlow to assess neural network resilience. Fault injections are applied through two primary methods: Layer State injection, targeting internal network components such as weights and biases, and Layer Output injection, which modifies layer outputs across various activations. Fault types include zeros, random values, and bit-flip operations, applied at varying levels and across different network layers. Our findings reveal several critical insights, such as the significance of bit-flip errors in critical bits, that can lead to substantial performance degradation or even system failure. With this work, we aim to exhaustively study the resilience of Real NVP models against errors due to radiation, providing a means to guide the implementation of fault tolerance measures.
\end{abstract}


\section{\textbf{Introduction}}
\begin{figure}[t] 
    \centering
    \includegraphics[width=\columnwidth]{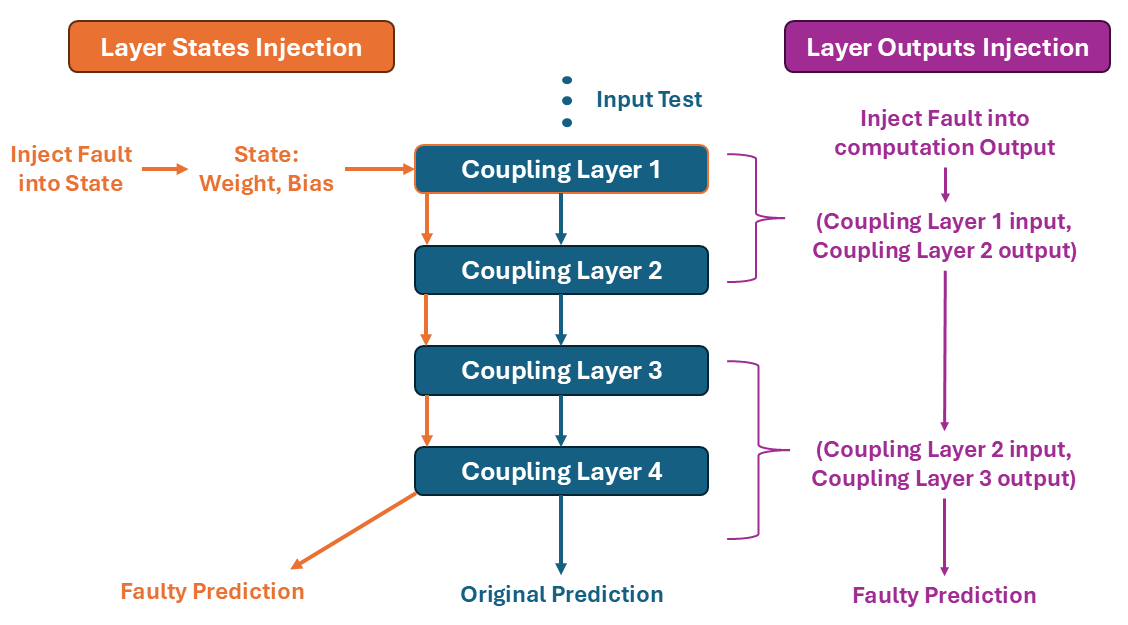}
    \captionsetup{justification=centering}
    \caption{Representation of Layer States and Outputs injections inside a Real NVP Network.}
    \label{injectionrepresentation}
\end{figure}

Satellites are essential for fields like communications, navigation, and Earth observation, as demonstrated by numerous missions, such as \cite{bepicolombo, argomoon, riccobono2022microsatellite}. As the number of space-based systems grows, so does the need to protect them from faults that can occur during their lifespan and from the harsh radiation environment, requiring robust hardware and software models. Neural Networks (NNs), known for handling large datasets and generating actionable insights, are increasingly being deployed in space missions for this purpose. For instance, ESA’s BepiColombo \cite{bepicolombo} leverages NNs in fault detection to maintain spacecraft health on its journey to Mercury. Among these methodologies, the Physics-Informed Real NVP model introduced by \cite{aim2024} demonstrated its ability to model complex distributions while incorporating domain-specific knowledge in the context of fault detection for an aerospace dataset. However, the complexity of NNs poses challenges for reliability assessment, especially under radiation conditions, as seen in Deep Space 1 (DS-1) \cite{paper_41}. Radiation effects are categorized as cumulative or single-event errors, we will focus on the latter. These are called Single-Event Effects (SEEs), and can be permanent or transient, such as soft errors like Single-Event Upsets (SEUs) and Single-Event Transients, both impacting system reliability. In Deep Neural Networks (DNNs), faults from radiation can alter outputs unpredictably, leading to Silent Data Corruption (SDC) or Detected Unrecoverable Errors (DUE). Although SDCs not affecting detection accuracy may be tolerable, those causing misdetection are critical.
Various methodologies exist for evaluating the reliability of computing devices, spanning different abstraction levels. Field Test involves exposing devices to natural particle flux and counting observed errors \cite{paper_35}. Beam Experiment induces faults by interacting accelerated particles with the silicon lattice at the transistor level, providing highly realistic error rates \cite{paper_8}. Microarchitecture-level fault injection offers broader fault coverage compared to software-level injection, as faults can potentially be injected across most system modules \cite{paper_36}. Software fault injection, performed at the highest level of abstraction, has proven effective in identifying code sections that are more susceptible to computational impact when corrupted \cite{paper_21}. Circuit- or gate-level simulations operate at the lowest level of abstraction, inducing either analog current spikes or digital faults while tracking fault propagation \cite{paper_37}. Finally, to enhance efficiency and maintain accuracy, hybrid approaches combining different abstraction levels are often employed \cite{paper_20}.

This study prioritizes understanding the impact of soft errors on NNs, through software fault injection, emphasizing the influence of model architecture and resilience rather than focusing on hardware-specific cumulative degradation. The fault injection framework requires careful engineering to avoid unrealistic results, because when evaluating the reliability of complex computing devices executing DNNs, it's crucial to consider that radiation-induced faults originate at the physical transistor level and then propagate through the architecture, ultimately affecting the software and modifying the output. Evaluations closer to the physical layer offer a more realistic perspective, while those closer to the software layer are more efficient \cite{paper_12}. In fact, while offering advantages such as lower costs, better controllability, and easier deployment for developers, the commonly adopted fault model (typically bit flip) 
\cite{paper_26, paper_29, paper_31, paper_38} may be accurate for main memory structures but less realistic for faults within computing cores or control logic, where the programmer has limited influence. 
Our approach involves emulating the effects of radiation by modifying memory values and altering key network elements, temporarily disabling them to assess their impact on performance and overall network behavior.
To sum-up, we propose a framework for testing Normalizing Flow networks, in particular Real NVP models, under simulated radiation-induced faults, to evaluate model robustness (see Figure \ref{injectionrepresentation}). It enables simultaneous testing of multiple models with varied hyper-parameters under different fault conditions, offering insights into model resilience.

\section{\textbf{Related Works}}
The resilience and robustness of DNNs have been widely studied by researchers \cite{paper_31, paper_32}. Before exploring various fault injection frameworks, it is crucial to understand the consequences of soft errors and the impact of perturbations introduced at different components of the network. \cite{paper_31} described the four main parameters that influence the impact of soft errors in DNNs:
\begin{enumerate}
    \item Topology and Data Type: Each DNN has a unique architecture and different combination of data types, both factors affect error propagation.
    \item Bit Position and Value: The sensitivity of each bit position varies, depending on the data type. High-order exponent bits are more likely to cause SDCs when corrupted, while mantissa and sign bits are less critical. Bit-flips from 0 to 1 in higher-order bits are more likely to cause errors than those from 1 to 0, as correct values in DNNs tend to cluster around zero \cite{paper_18}.
    \item Layers: Errors propagate in distinct ways across different layers depending on their type and position.
    \item Data Reuse: Data reuse strategies in DNN accelerators’ dataflows impact the Silent Data Corruption (SDC) probability.
\end{enumerate}
The authors of \cite{paper_31} reach these results by testing several networks, such as AlexNet \cite{alexnet}, CaffeNet \cite{caffenet}, and NiN \cite{ninnet}, using a DNN simulator on various datasets to inject faults and evaluate performance by calculating the SDC Probability and the Failure-in-Time (FIT) rate.
When considering layer position and type, \cite{paper_31} has highlighted the non-uniform impact of faults on layers positioned at different depths. Faults occurring in earlier layers are more likely to propagate, but many are masked by operations like pooling or ReLU. Additionally, the numerical magnitude of faulty activations, rather than their quantity, significantly influences the probability of SDC errors. Finally, the above-mentioned paper suggests strategies to improve resilience: (I) DNNs should use a type that offers sufficient numerical range and precision to operate with success, as restricting data types and suppressing dynamic value ranges can help mitigate the effects of bit-flips; (II) normalization layers can improve accuracy by mitigating SDCs, as faulty values are normalized alongside fault-free values. While these insights provide a foundation, they primarily focus on DNNs and may not fully capture the nuances of time-series data. To address this gap, our contribution delves into the resilience of 32-bit floating-point precision Real NVP models designed for high-dimensional time-series data.

\section{\textbf{Methodology}}
In this section we provide an overview of the neural network architecture and loss, and of the metric used, and describe the considered dataset.

\subsection{\textbf{Physics-Informed Real NVP}}
Real NVP \cite{realnvp_paper} is a neural network architecture belonging to the class of normalizing flows models, a class of generative models, that transforms a simple probability distribution into a more complex one that matches the data distribution through a sequence of invertible functions. Real NVP introduces non-volume-preserving transformations, which offer more flexibility in modeling complex distributions while still allowing efficient computation of the determinant of the Jacobian, which is necessary for calculating densities. Coupling Layers are the fundamental component of these networks, built as a flexible and tractable bijective function composed of two Fully-Connected (FC) neural networks that use half of the input, which remains unaltered, to compute respectively a scale and a translation factor to be applied to the remaining half of the input. \cite{aim2024} enhanced this architecture with a physics-informed loss created by extracting relationships from the considered dataset, ADAPT \cite{adapt_paper}.

\subsection{\textbf{Dataset}}
We used the same dataset and configuration used by \cite{aim2024}. The training and testing phases of the NN models are performed with the ADAPT dataset \cite{adapt_paper}, an Electrical Power System (EPS) dataset. In particular, given the dataset, we created 3 splits. This allows us to train three models with the same hyper-parameters on three distinct training sets, which are then evaluated on their corresponding test sets. The final score for a specific metric is calculated as the average of the scores obtained from each model. Therefore, when we refer to the results of a model, we mean the average across these different splits.

\begin{figure}[t]
    \centering
    \begin{subfigure}[b]{0.24\textwidth}
        \centering
        \includegraphics[width=1\textwidth]{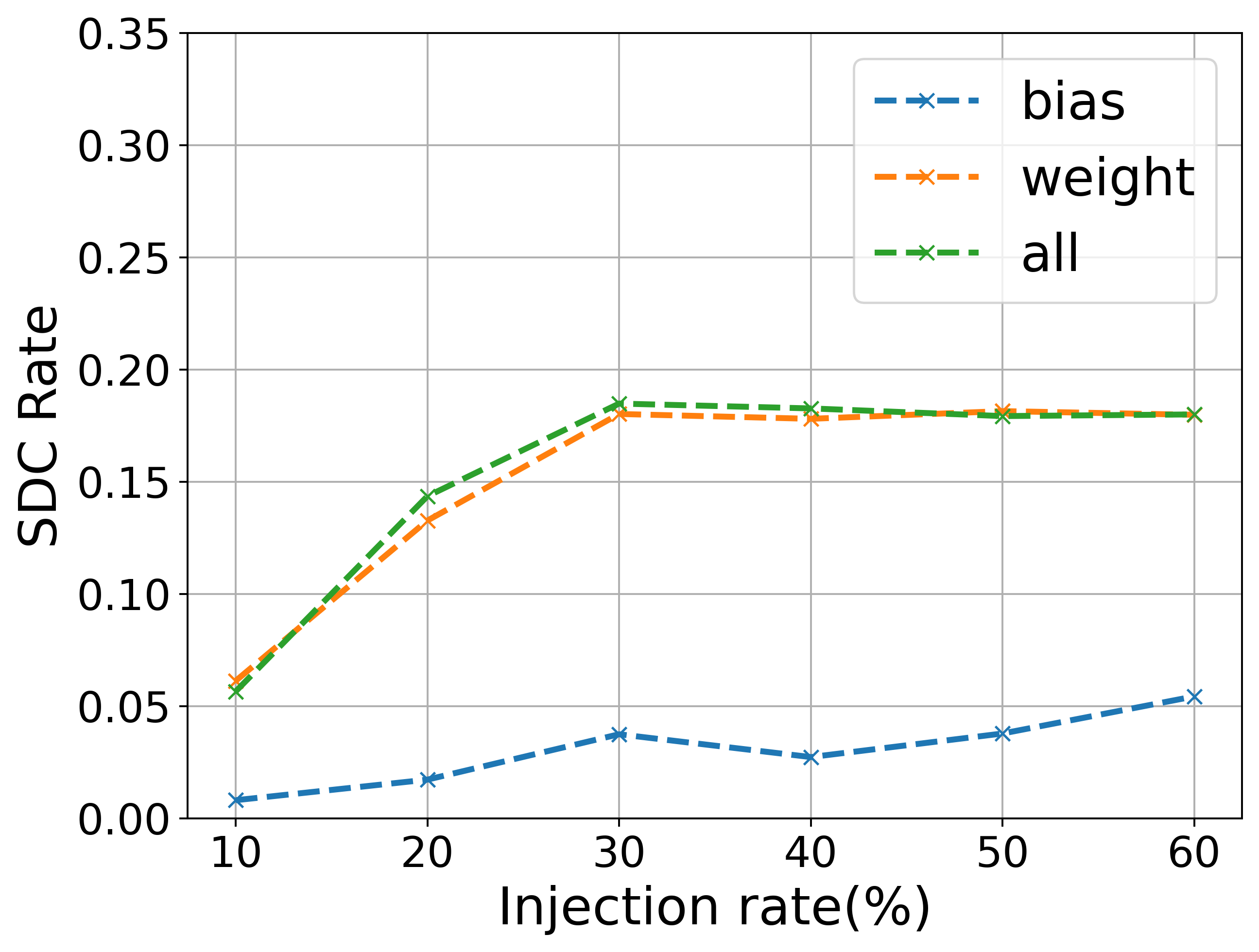}
        \captionsetup{font=footnotesize, justification=centering}
        \caption{Zeros injection mode=100\%.}
        \label{stateszeros}
    \end{subfigure}
    \begin{subfigure}[b]{0.24\textwidth}
        \centering
        \includegraphics[width=1\textwidth]{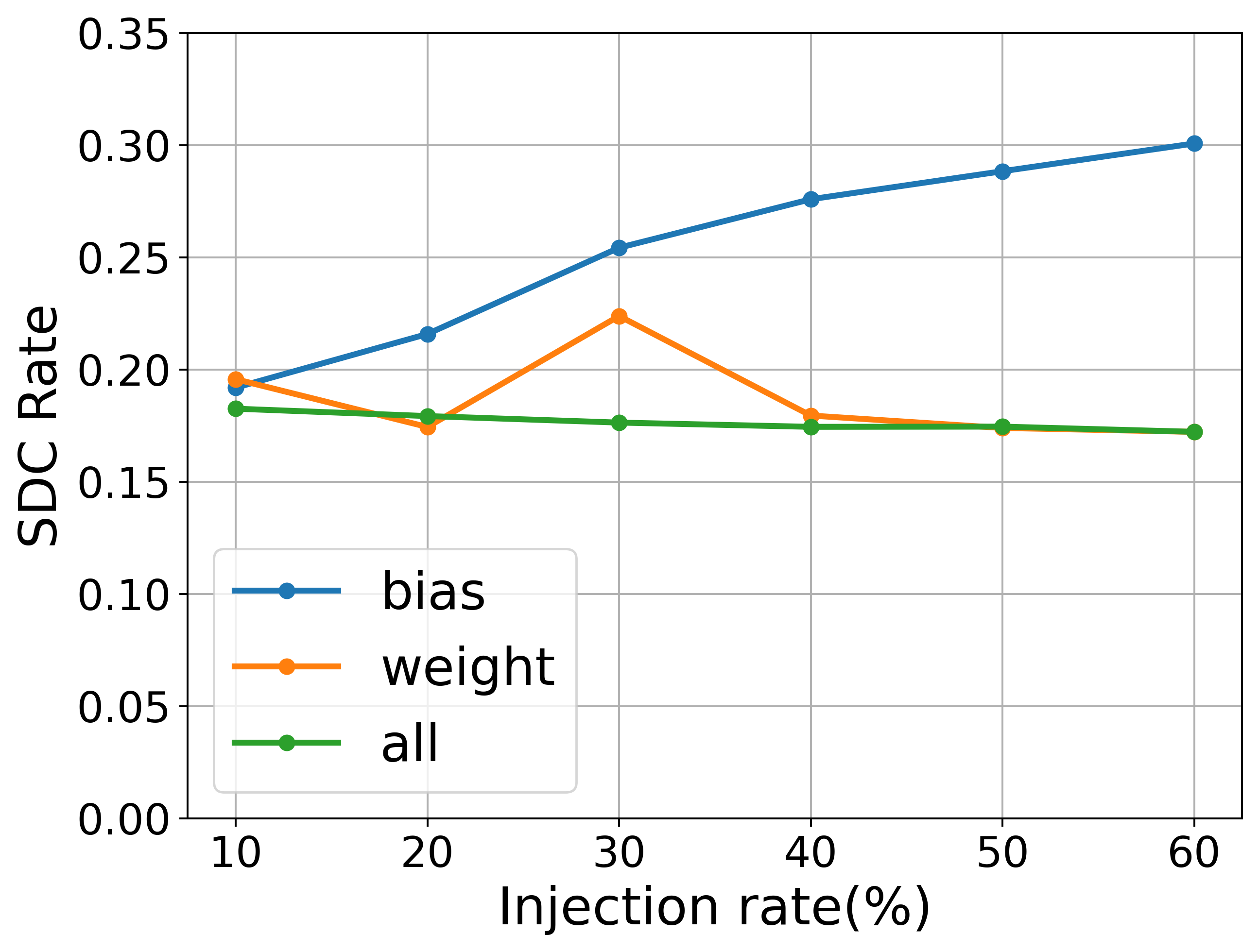}
        \captionsetup{font=footnotesize, justification=centering}
        \caption{Random injection mode=100\%.}
        \label{statesrandom}
    \end{subfigure}
    \begin{subfigure}[b]{0.24\textwidth}
        \centering
        \includegraphics[width=1\textwidth]{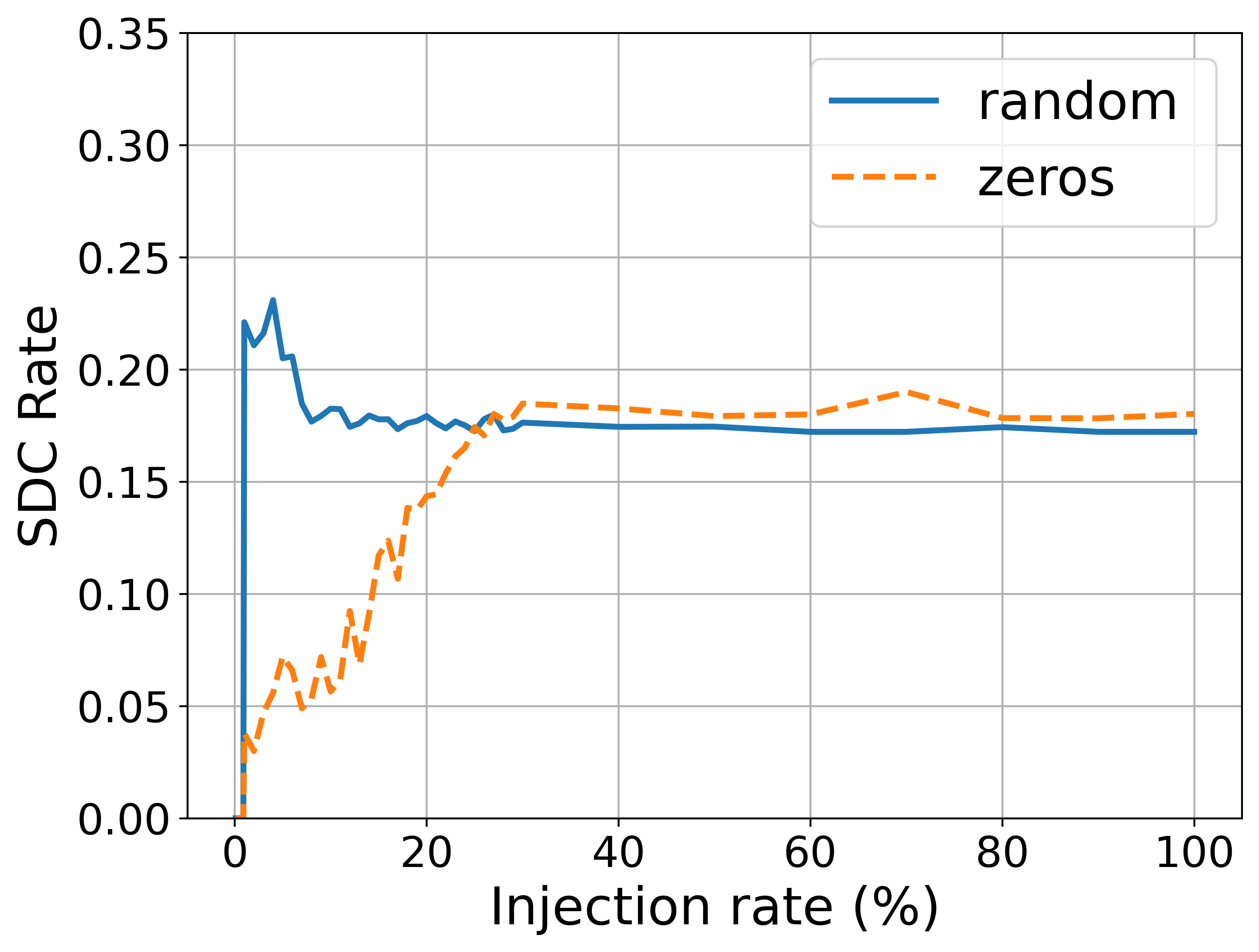}
        \captionsetup{font=footnotesize, justification=centering}
        \caption{Injections from 0 to 100 layers percentage.}
        \label{layer_states_saturation}
    \end{subfigure}
    \begin{subfigure}[b]{0.24\textwidth}
        \centering
        \includegraphics[width=1\textwidth]{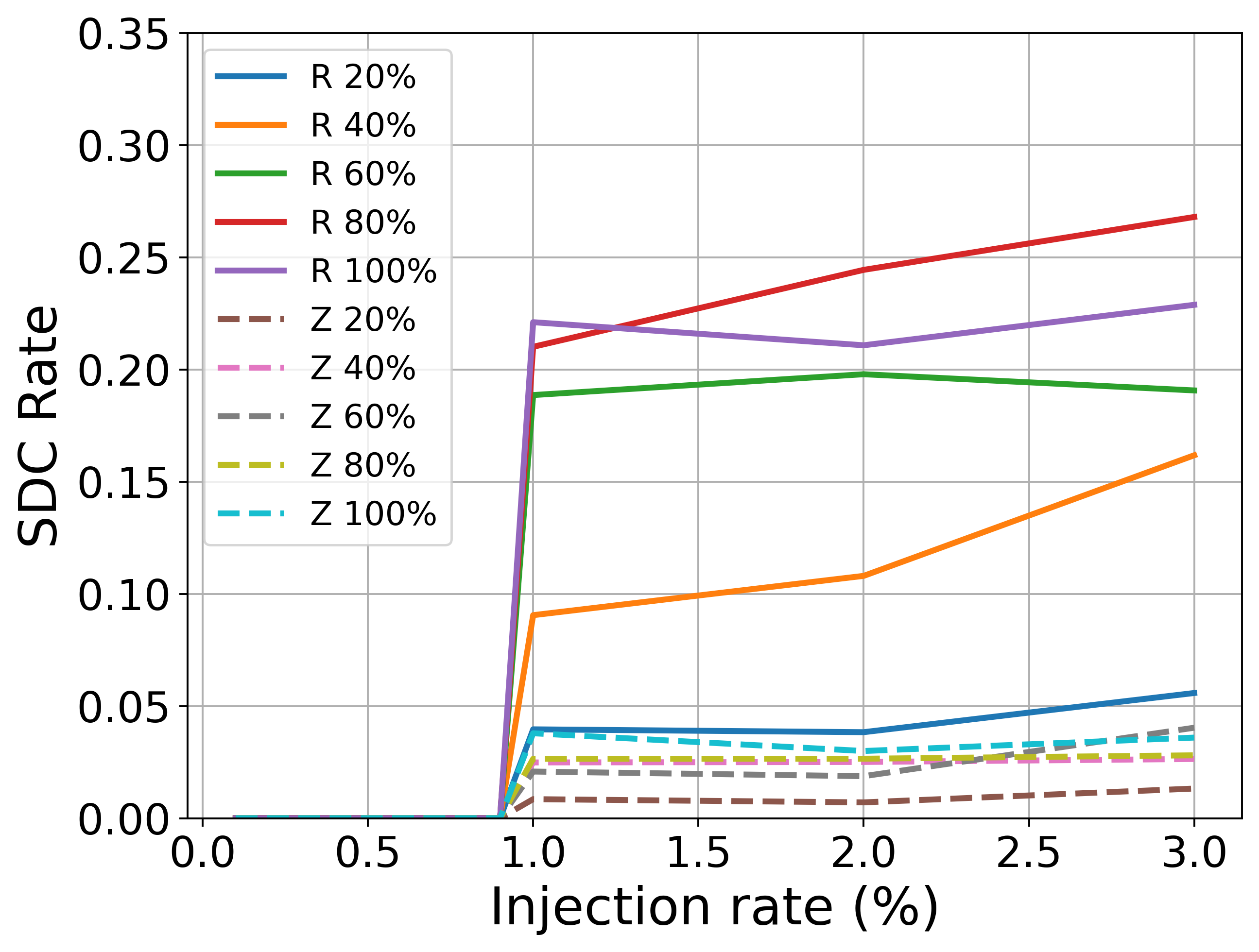}
        \captionsetup{font=footnotesize, justification=centering}
        \caption{Start of performance degradation for Zeros (Z) and Random (R).}
        \label{stateszerosrandomstart}
    \end{subfigure}
    \captionsetup{justification=centering}
    \caption{Zeros and Random Layer States injections.}
    \label{layer_states_zeros_random}
\end{figure}

\subsection{\textbf{Fault Injections}}
\label{sec:fault_inj}
This section explores the layer states and layer outputs injection processes (Figure \ref{injectionrepresentation}) to assess the impact of targeted modifications on neural network behavior. These processes are controlled by several parameters, allowing for a wide range of configurations. The parameters used to evaluate the model behavior when injecting the layer states are:
\begin{itemize}
    \item \textbf{Mode}: Specifies the percentage of layers to be injected, with options ranging from 20\% to 100\%.
    \item \textbf{Variable}: Determines the type of variable targeted between bias, weight, or both of them at the same time, defined as 'all' in the plots.
    \item \textbf{Type}: Defines the injection method by Zeros (sets variables to zero), Random (applies Gaussian noise), and Bit-flips (flips specific bits in the values).
    \item \textbf{Amount}: Sets the injection rate for each layer, expressed as a percentage.
    \item \textbf{Bit}: Sets a specific bit position to be flipped or a random one.
    \item \textbf{Direction}: Used only for bit-flips, it controls the flipping direction (0 to 1, 1 to 0, or both).
    \item \textbf{Sign}: Limits bit-flips applied to the sign bit to positive, negative, or both types of values.
\end{itemize}
In the Layer Outputs experiments, we used several of the above-mentioned parameters (Type, Amount, Bit, Direction, and Sign) along with:
\begin{itemize}
    \item \textbf{Mode}: Specifies the coupling layer to inject, either as a specific layer, a random layer, or all layers.
    \item \textbf{Variable}: Targets either "scale", "translation", or both types at the same time.
    \item \textbf{Activation}: Chooses the activation functions to be used as targets for injection among ReLU, used in all hidden layers, Tanh or Linear, used respectively in the last layers of the scale and translation networks, or all of them, referenced as "all" in the plots.
    \item \textbf{Method}: Given one or more coupling layers, it determines the injection location by picking "Partial" (final FC layers) or "Complete" (all FC layers).
\end{itemize}

The bit-flip operation pertains to the manipulation of individual bits in the IEEE 754 \cite{floatingpoint} single-precision floating-point format (binary32), a widely used 32-bit representation for numerical values. This format consists of a 1-bit sign (S), an 8-bit exponent (E), and a 23-bit significand (F) (with an implicit leading 1), enabling a dynamic range of values defined by (3).
\begin{equation}
    VALUE = (-1)^S\cdot 2^{(E-127)}\cdot(1.F)
\end{equation}
This structure ensures approximately 6 to 9 significant decimal digits of precision. Special bit patterns in the exponent define unique cases, such as de-normalized numbers or zero (E = 0) and infinity/NaN (E = 255), which are critical in numerical computations.
Fault injections in layer states enable us to evaluate the model's ability to handle corrupted weights and biases. Injection in layer outputs reveals differences in the resilience of scale and translation operations.

\subsection{\textbf{Metric}}
To assess model robustness under fault injection, we used a resilience metric, the SDC rate, which evaluates the fraction of injected faults leading to silent data corruptions. The SDC rate is defined in (1).
\begin{equation}
    SDC_{rate} = \frac{\sum_{}^{N_{exps}}\sum_{}^{N_{seeds}} SDC_{rate\_exp}}{N_{exps} \cdot N_{seeds}}
\end{equation}
Where $N_{seeds}$ is the number of different seeds, $N_{exps}$ is the number of experiments executed for each seed, and $SDC_{rate\_exp}$ is the SDC rate of each experiment, defined in (2).
\begin{equation}
    SDC_{rate\_exp} = \frac{1}{N_{samples}}\sum_{i=1}^{N_{samples}}\begin{cases}1\text{, if $\textbf{x}_i$ misclassified}\\ 0\text{, otherwise}\end{cases}
\end{equation}
Where $\textbf{x}$ is the vector of $N_{samples}$ samples correctly classified by the Real NVP when no faults were injected. By selectively injecting faults at specific network layers or activation functions, our approach offers both model-level and variable-level insights into the resilience of machine learning models, aligning with metrics like PVF \cite{pvfvulnerabilityfactor}  to enhance our understanding of model reliability. The SDC rate is calculated using two approaches: the absolute SDC metric, which evaluates all models against a fixed set of correctly classified test samples, and the relative SDC rate, computed individually for each model based on its unique accuracy and test subset. The absolute SDC provides consistent comparisons across models, while the relative SDC accounts for model-specific performance variations.

\section{\textbf{Experimental Results}}
\subsection{\textbf{Setup}}
\label{ls_setup}
Given a pre-trained model, we performed multiple experiments based on the parameter $N_{exps}$, which we set equal to 10 to ensure a reliable SDC rate, with $N_{seeds}=3$ random seeds. For each experiment, the NN state was reset to its original state after every iteration, and the scores for each metric were averaged across all experiments and seeds. For the single-model analysis, presented in sections \ref{zerosrandomsection} and \ref{bitflipssection}, we used the smallest model, configured with 4 coupling layers with 3 FC layers each, and 32 units per FC layer. Bit-flip injections in the layer states were performed with a fixed injection rate of 10\%, targeting all layers ("Mode" = 100\%). For the multiple-model analysis, presented in section \ref{multiplemodels}, we evaluated 18 different models trained using various hyper-parameter combinations. These combinations included the number of coupling layers (4 or 6), of FC layers for each coupling layer (3, 4, or 5), and the number of units in each FC layer (32, 48, or 64).

\subsection{\textbf{Zeros \& Random}}
\label{zerosrandomsection}

\begin{figure}[t] 
    \centering
    \includegraphics[width=1\columnwidth]{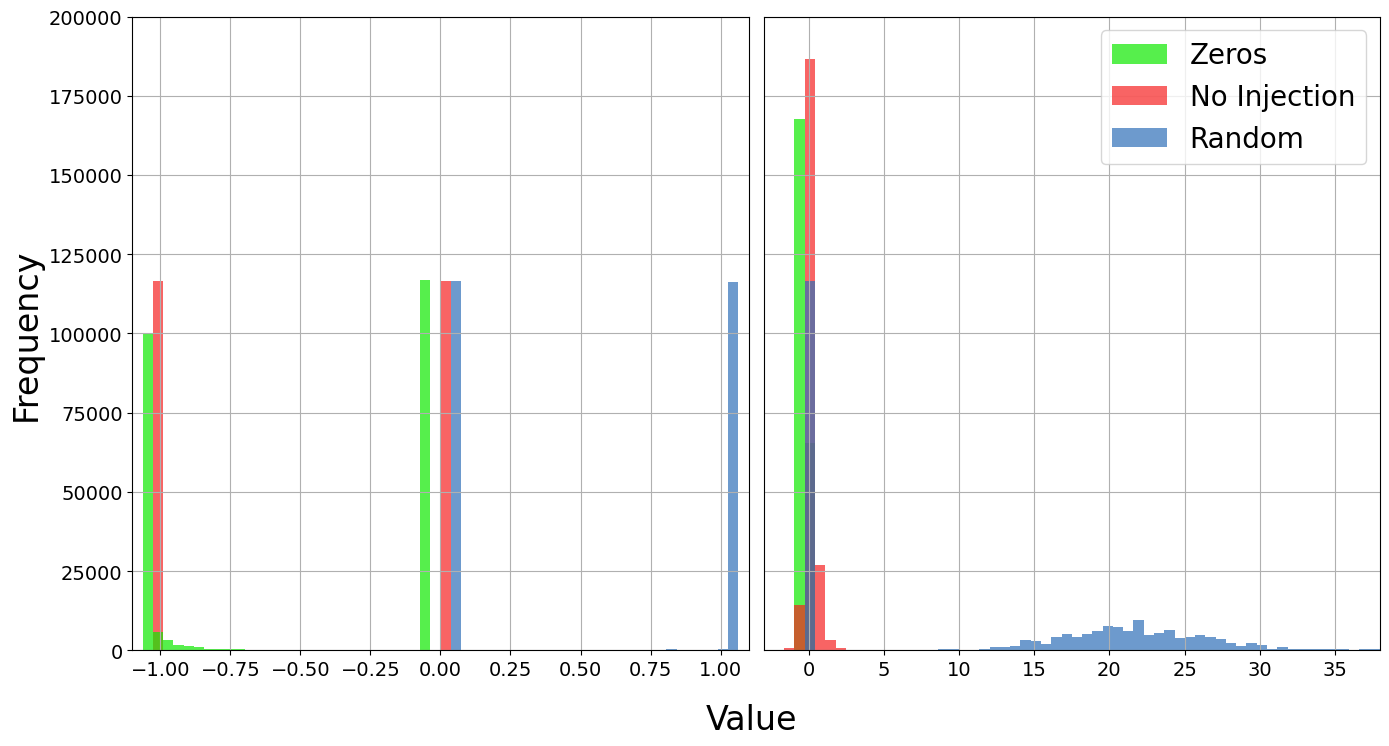}
    \captionsetup{justification=centering}
    \caption{Distribution of scale (Right) and translation (Left) masked output of the last layer of the network on three different injection configurations.}
    \label{output_from_layer}
\end{figure}

\textbf{Layer States:} Figure \ref{layer_states_zeros_random} reveals distinct effects of fault injections on the Bias and Weight variables of the Real NVP network, collectively referred to as "all". When injecting faults into all layers, the SDC rate differs significantly depending on the injection type and the variables targeted. Injecting zero values into the bias results in minor performance degradation. Removing it does not disrupt the relative weighting of the layers' outputs, as it primarily adjusts their shift. In contrast, injecting random noise into the bias causes greater degradation, introducing global perturbations to the output and disrupting the controlled behavior of the activation functions. Injecting zeros into the weights causes significant degradation, as it effectively cuts off specific connections within the network, leading to a collapse in learned representations by pruning parts of the network. On the other hand, injecting random noise into the weights has a less drastic impact. The redundancy and distributed nature of weight connections mean that noise often affects only a subset of weights, allowing the model to mitigate the disruption through averaging effects. Performance degradation begins at an injection rate of approximately 0.8\% (Figure \ref{stateszerosrandomstart}), and as injection rates increase we observed a saturation in performance (Figure \ref{layer_states_saturation}): Zeros injections plateau around a 30\% injection rate, while Random injections saturate earlier, at approximately 10\%. This behavior stems from the network's internal architecture and the use of tanh activations in scale layers and linear activations in translation layers. Due to these factors, the model can rely on the input alone to drive the output, as the input influences the propagation and transformation of values through the network, even when all variables are zeroed or randomized. Detailed analysis in Figure \ref{output_from_layer} shows that:
\begin{itemize}
    \item \textbf{Zero injections} result in scale layer outputs oscillating between -1 and 0, with values saturating near -1 due to the tanh activation. Translation layers shift outputs to higher values, with ReLU modifying only negative inputs.
    \item \textbf{Random injections} confine masked scale outputs near 0 or 1, while translation layers produce higher deviations from expected outputs.
\end{itemize}
These behaviors, linked to activation functions and the propagation of masked operations, reveal critical insights into the network’s robustness.

\begin{figure}[t]
    \centering
    \begin{subfigure}[b]{0.24\textwidth}
        \centering
        \includegraphics[width=1\textwidth]{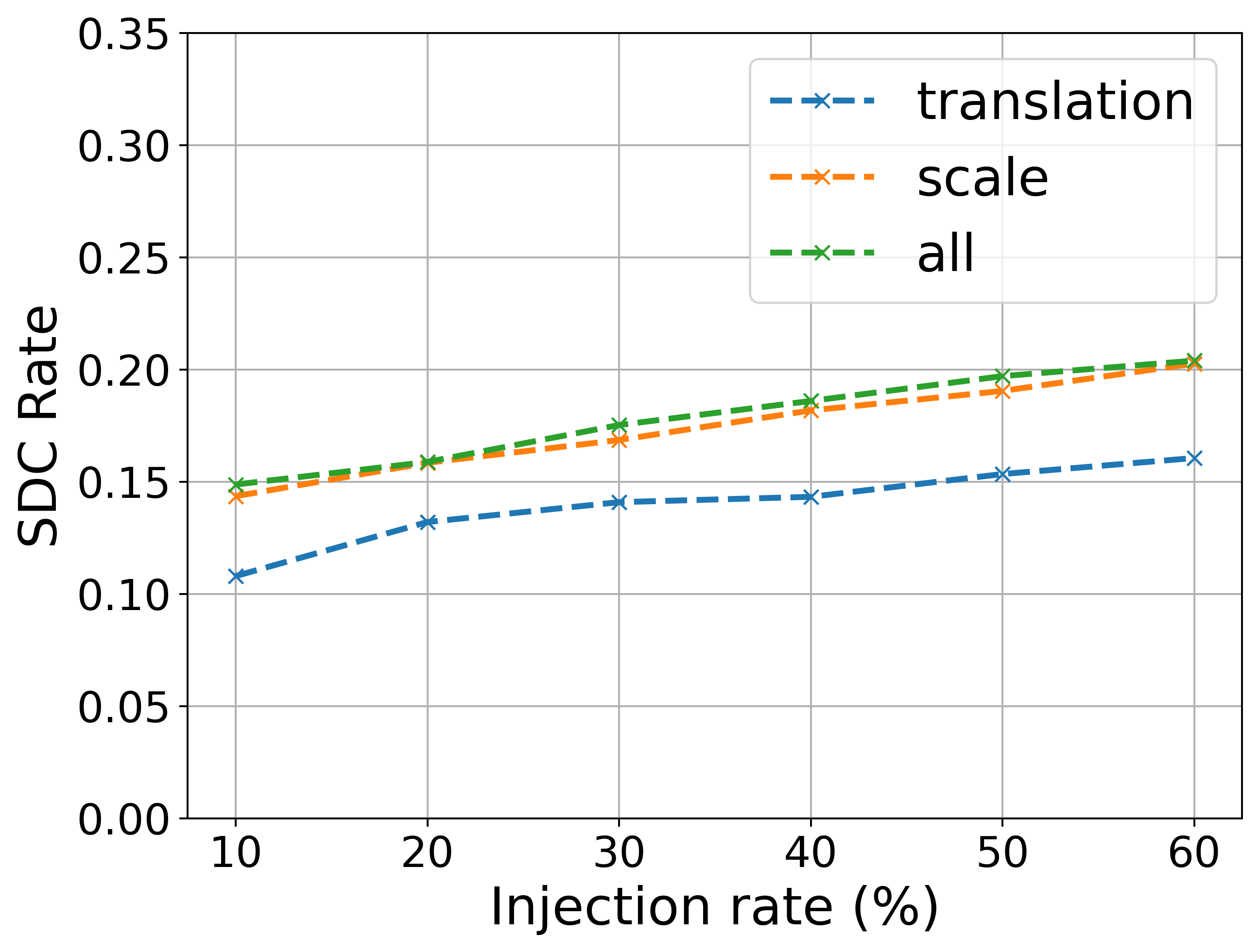}
        \captionsetup{font=footnotesize, justification=centering}
        \caption{Zeros injection "Partial".}
        \label{zerospartial}
    \end{subfigure}
    \begin{subfigure}[b]{0.24\textwidth}
        \centering
        \includegraphics[width=1\textwidth]{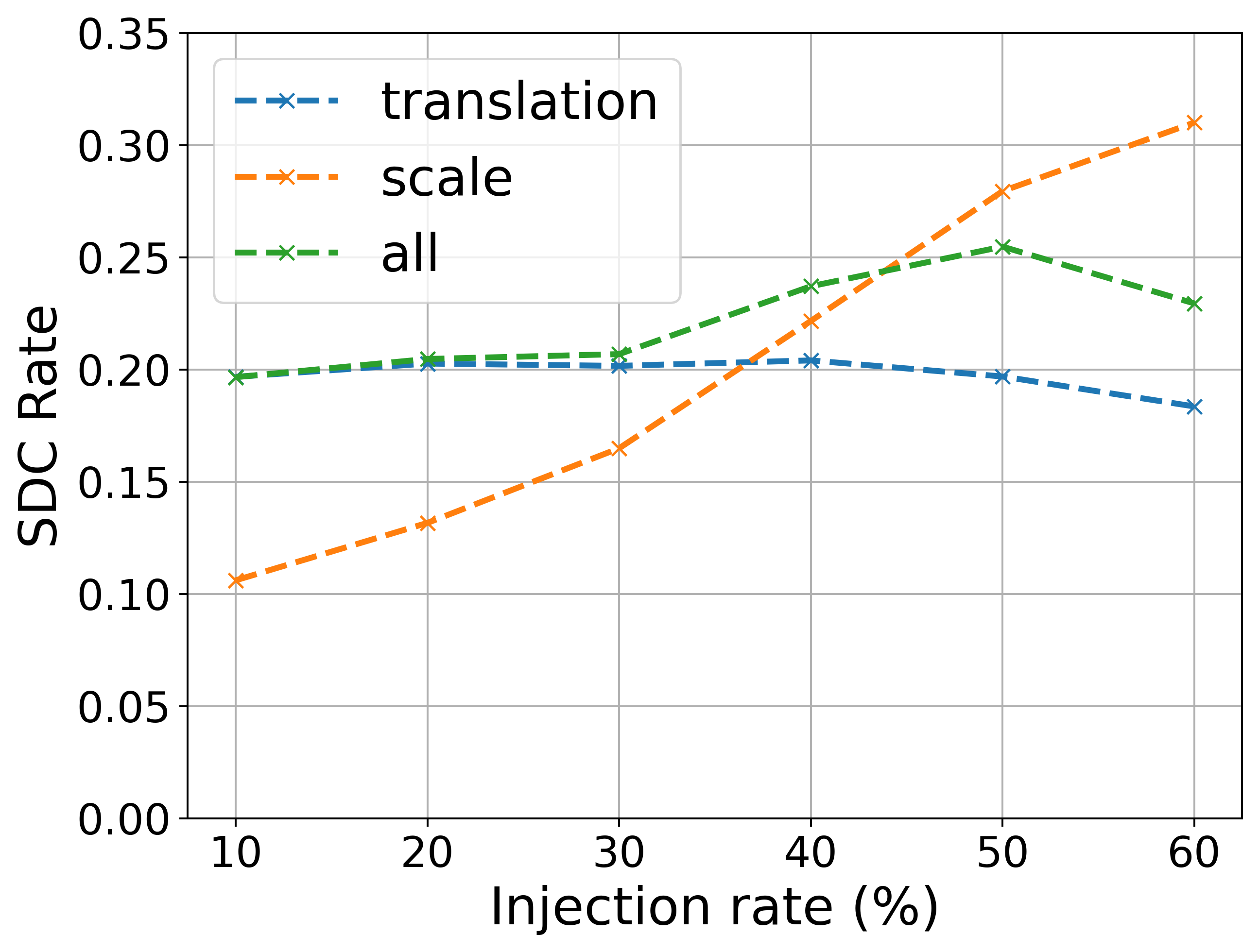}
        \captionsetup{font=footnotesize, justification=centering}
        \caption{Zeros injection "Complete".}
        \label{zeroscomplete}
    \end{subfigure}
    \begin{subfigure}[b]{0.24\textwidth}
        \centering
        \includegraphics[width=1\textwidth]{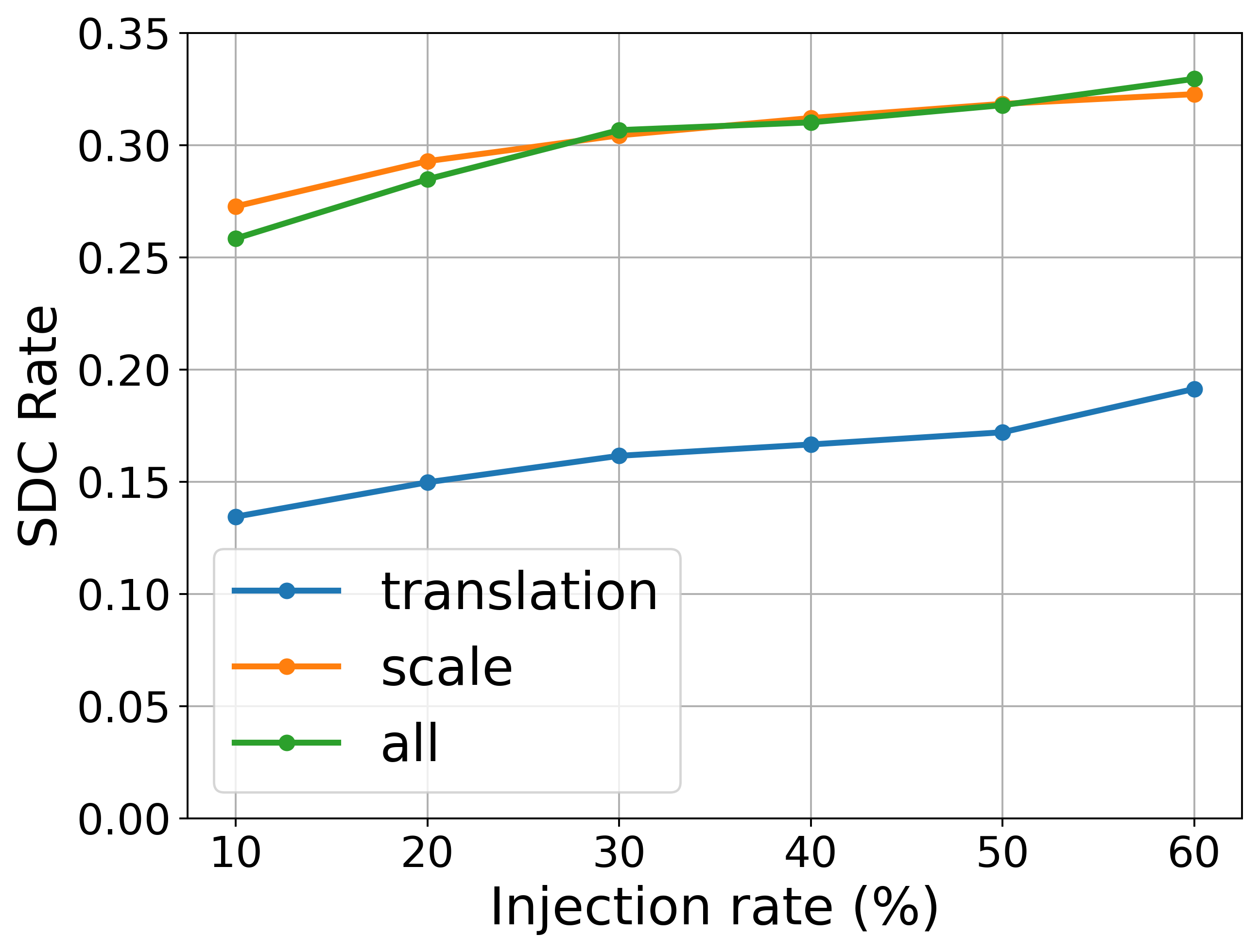}
        \captionsetup{font=footnotesize, justification=centering}
        \caption{Random injection "Partial".}
        \label{randompartial}
    \end{subfigure}
    \begin{subfigure}[b]{0.24\textwidth}
        \centering
        \includegraphics[width=1\textwidth]{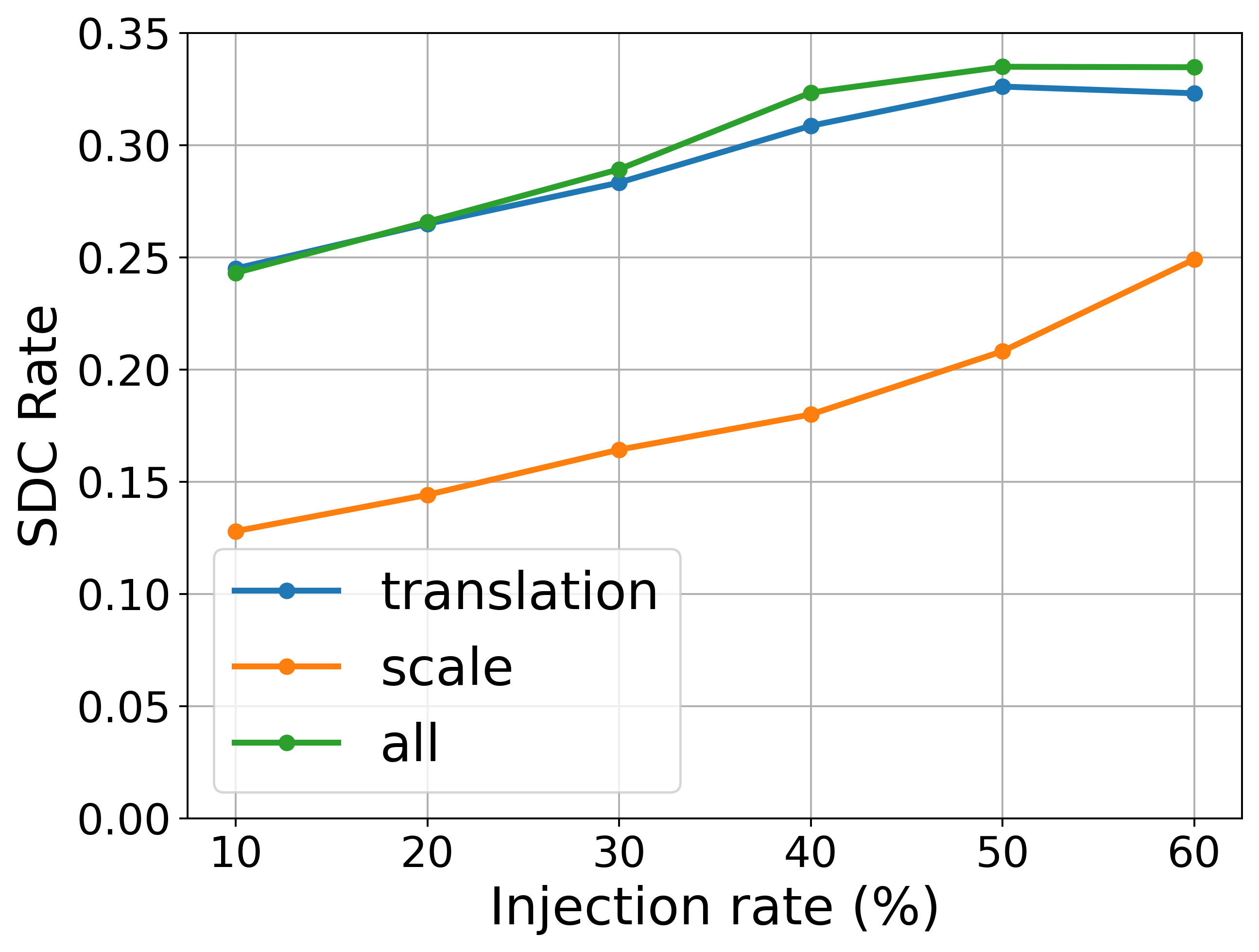}
        \captionsetup{font=footnotesize, justification=centering}
        \caption{Random injection "Complete".}
        \label{randomcomplete}
    \end{subfigure}
    \captionsetup{justification=centering}
    \caption{Zeros and Random Layer Outputs injections in all layers.}
    \label{layer_outputs_scale_translation}
\end{figure}

\begin{figure}[t]
    \centering
    \begin{subfigure}[b]{0.49\textwidth}
        \centering
        \includegraphics[width=0.9\textwidth]{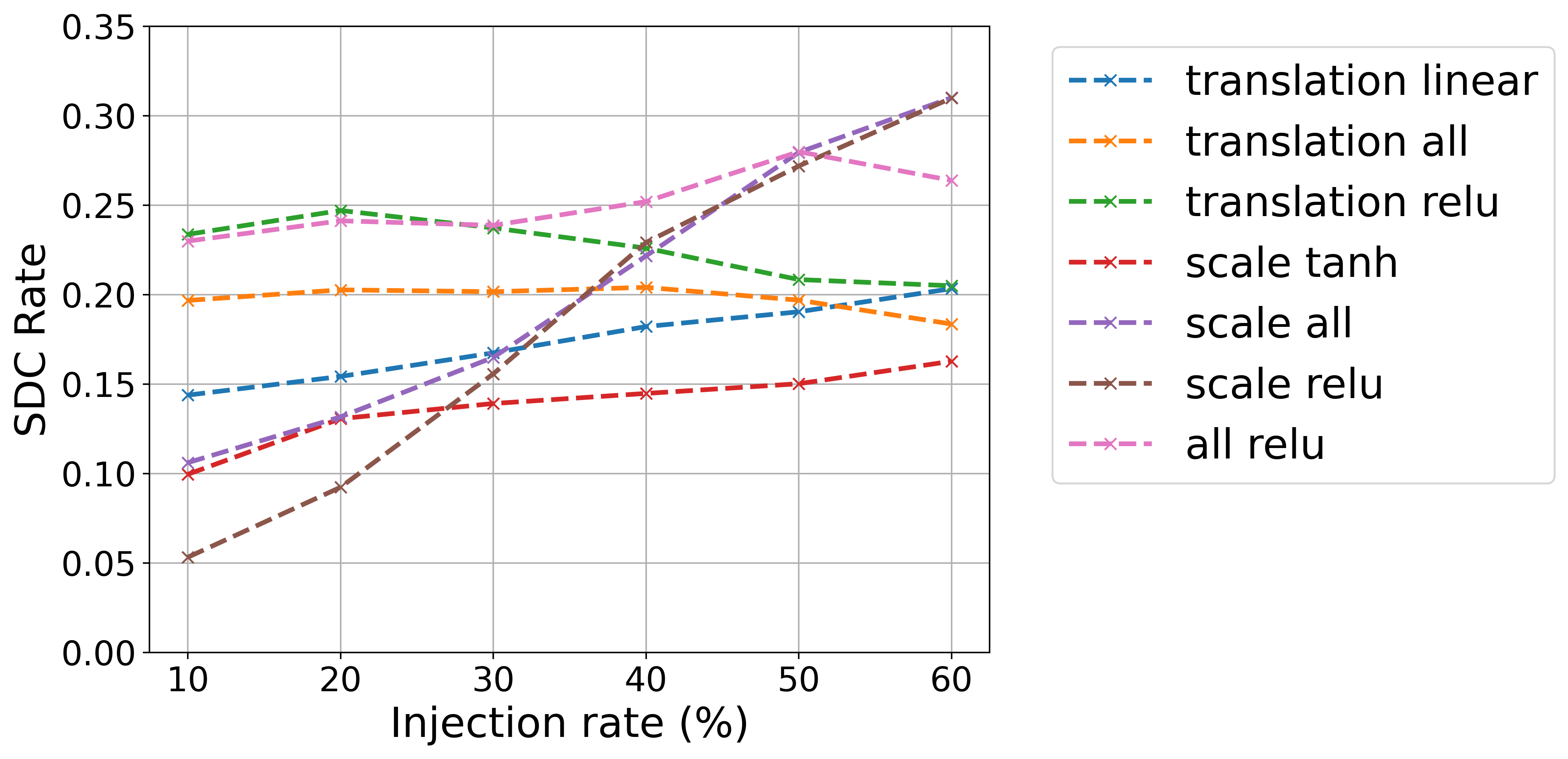}
        \captionsetup{justification=centering}
        \caption{Zeros injections at different activations in all layers.}
        \label{outputzerosactivation}
    \end{subfigure}
    \begin{subfigure}[b]{0.49\textwidth}
        \centering
        \includegraphics[width=0.9\textwidth]{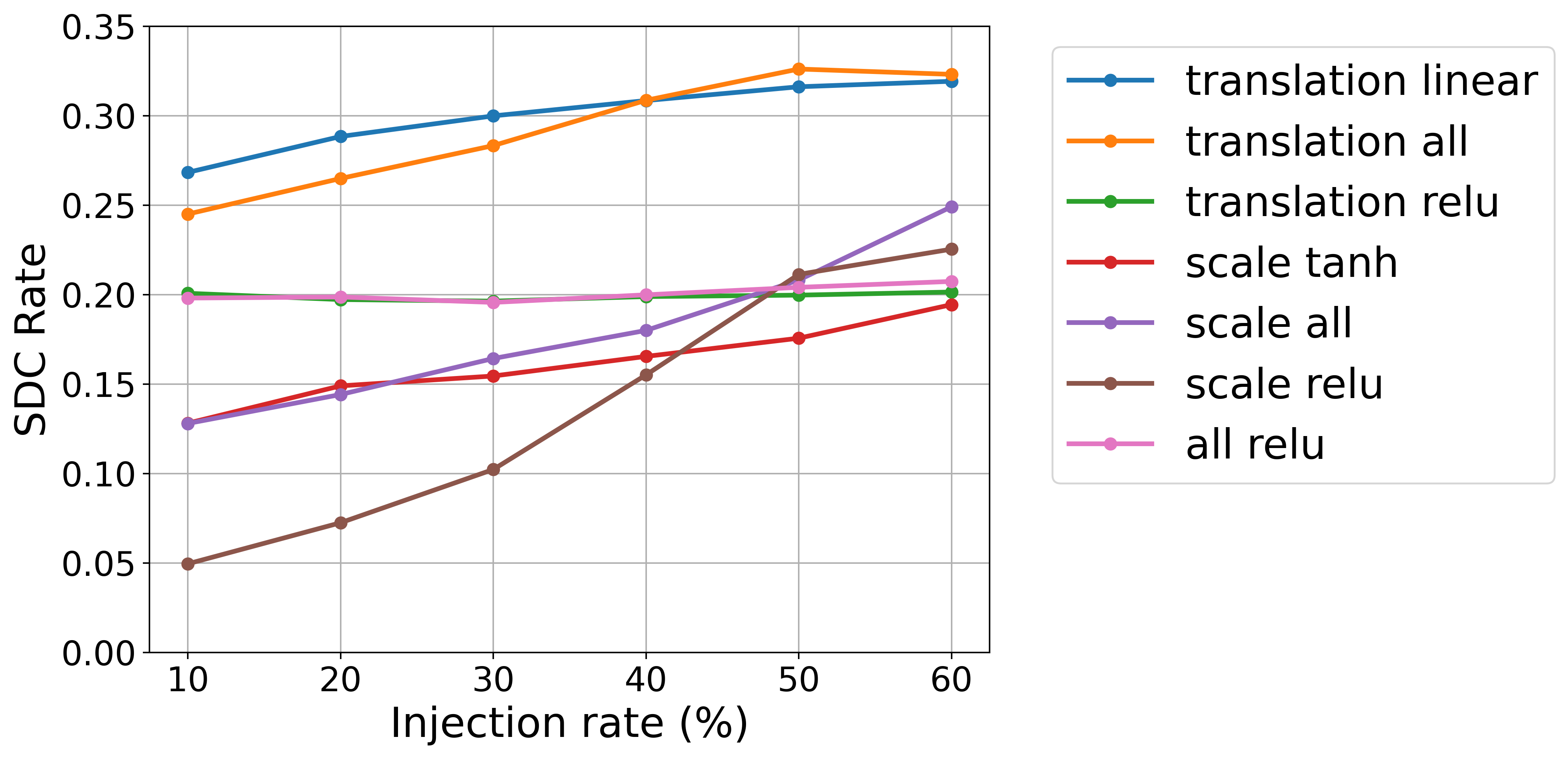}
        \captionsetup{justification=centering}
        \caption{Random injections at different activations in all layers.}
        \label{outputrandomactivation}
    \end{subfigure}
    \caption{Layer Outputs injections on activation functions.}
    \label{layer_outputs_activations_zeros_random}
\end{figure}

\begin{figure}[b]
    \centering
    \begin{subfigure}[b]{0.24\textwidth}
        \centering
        \includegraphics[width=1\textwidth]{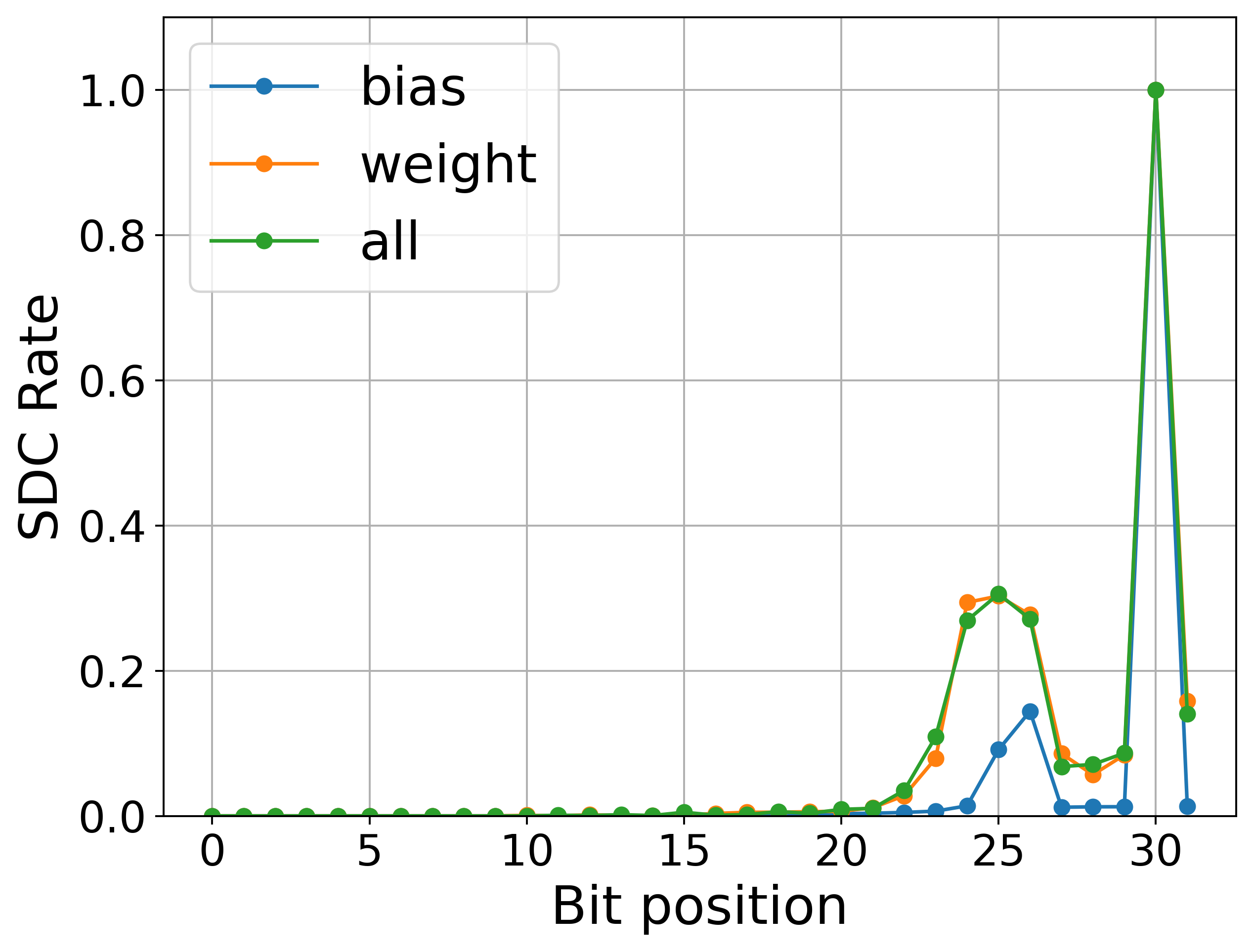}
        \captionsetup{font=footnotesize, justification=centering}
        \caption{Layer States injection with all directions and signs.}
        \label{statesbitflips}
    \end{subfigure}
    \begin{subfigure}[b]{0.24\textwidth}
        \centering
        \includegraphics[width=1\textwidth]{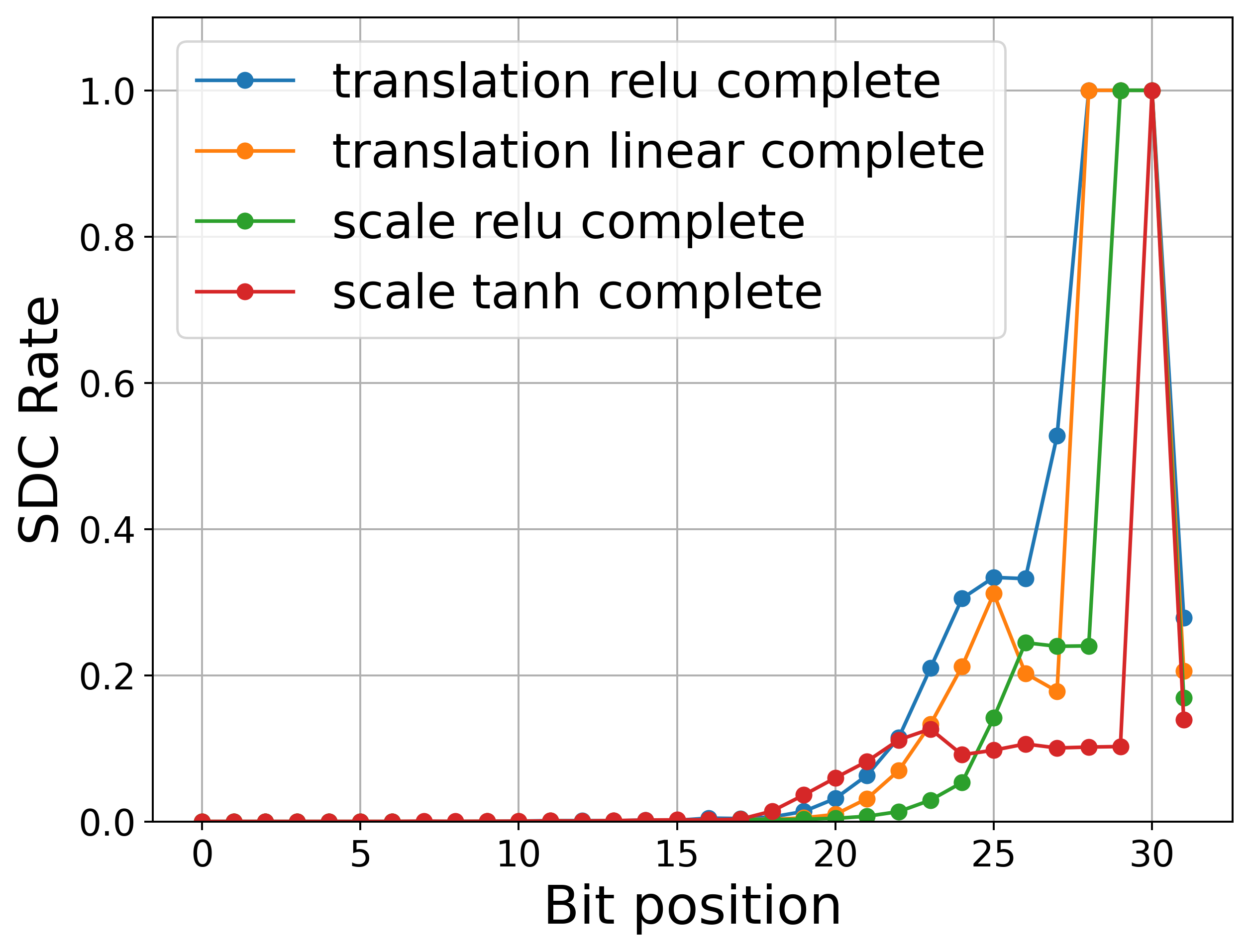}
        \captionsetup{font=footnotesize, justification=centering}
        \caption{Layer Outputs injection in all layers at different Activations.}
        \label{bitflipsactivations}
    \end{subfigure}
    \begin{subfigure}[b]{0.24\textwidth}
        \centering
        \includegraphics[width=1\textwidth]{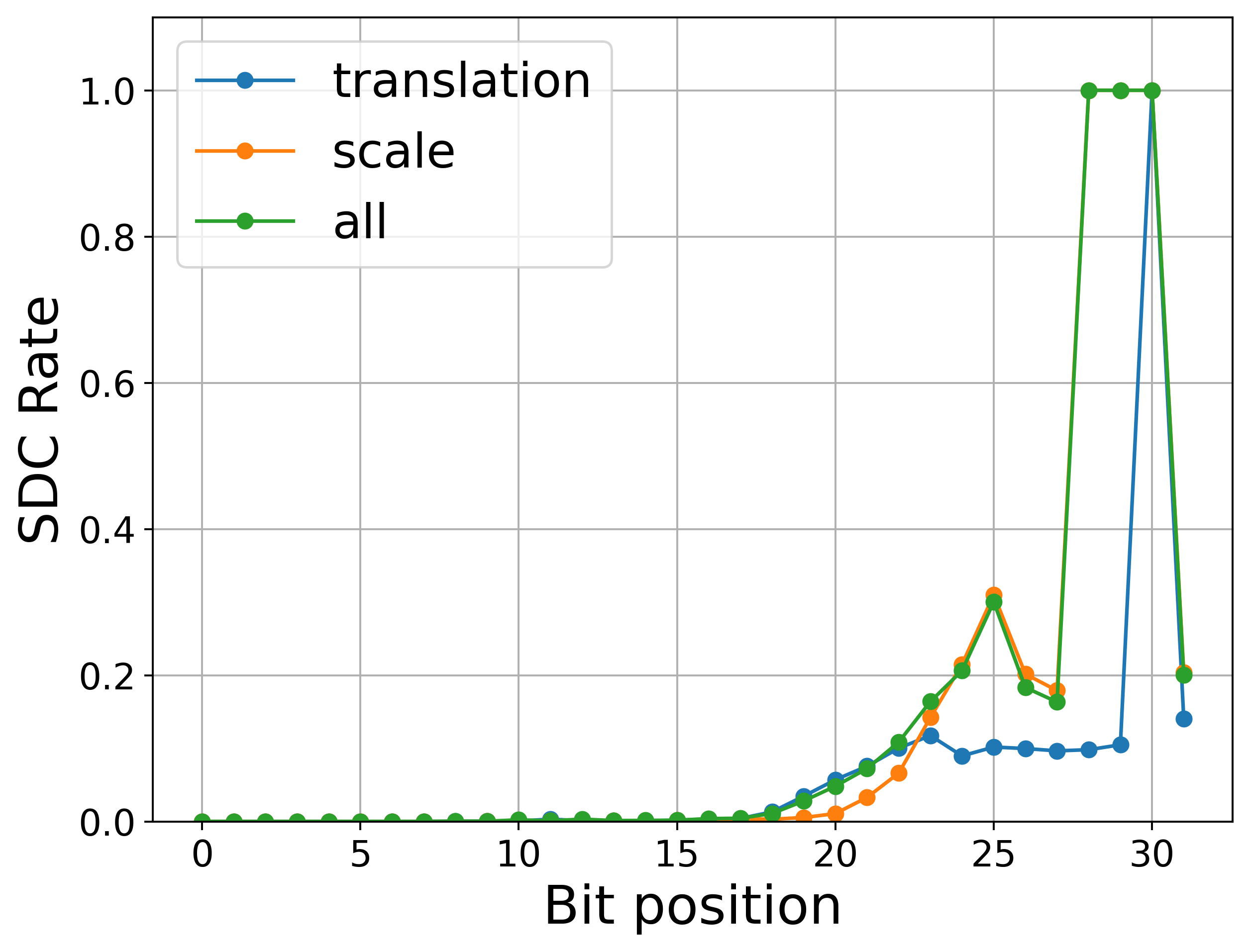}
        \captionsetup{font=footnotesize, justification=centering}
        \caption{Layer Outputs injection "Partial" in all layers.}
        \label{outputsbitflipspartial}
    \end{subfigure}
    \begin{subfigure}[b]{0.24\textwidth}
        \centering
        \includegraphics[width=1\textwidth]{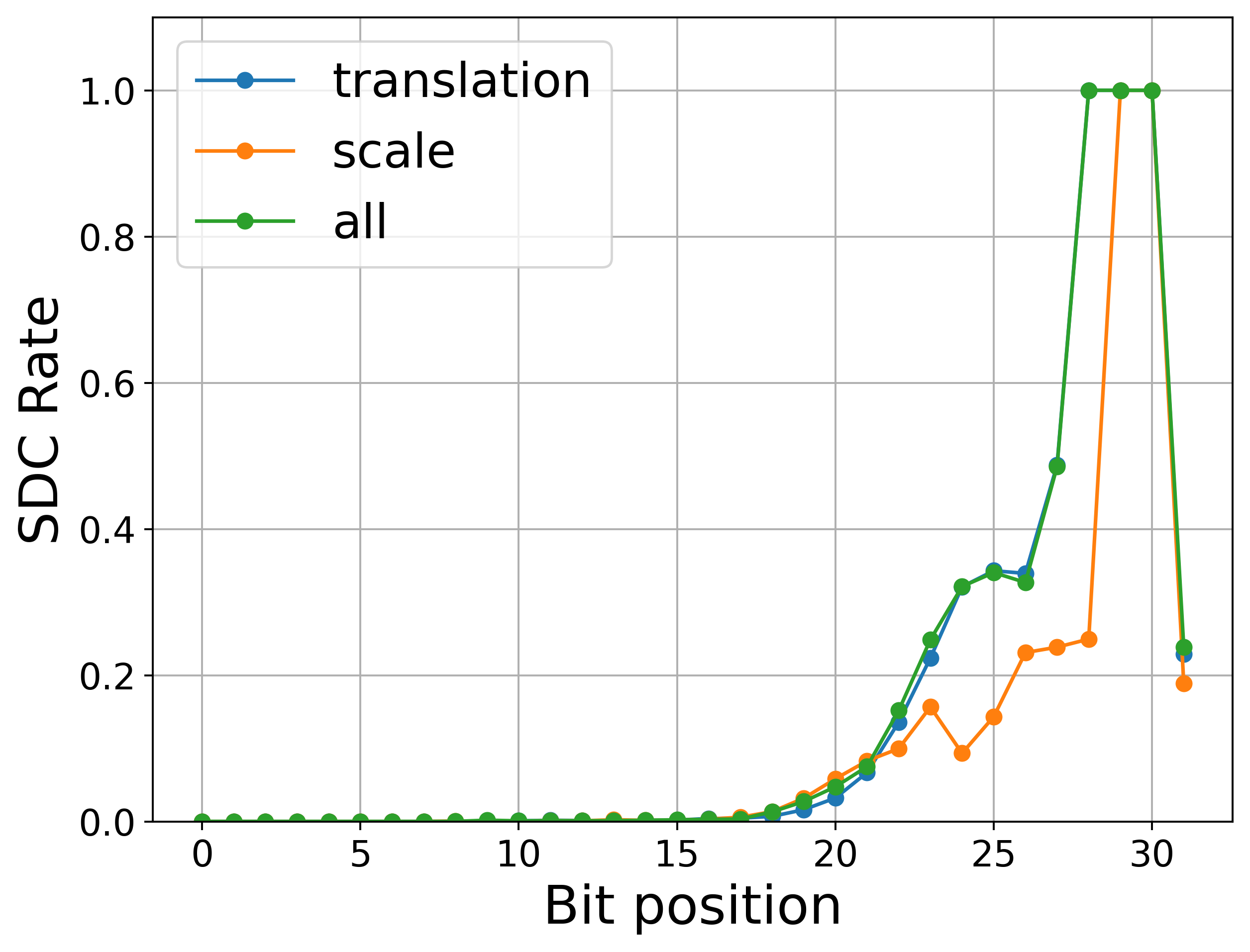}
        \captionsetup{font=footnotesize, justification=centering}
        \caption{Layer Outputs injection "Complete" in all layers.}
        \label{outputsbitflipscomplete}
    \end{subfigure}
    \caption{Bit-flips injections.}
    \label{layer_states_bitflips}
\end{figure}

\textbf{Layer Outputs:} This analysis examines the impact of fault injections on layer outputs, focusing on the internal network behavior with unaltered variables. In Figure \ref{layer_outputs_scale_translation} for "Partial" injections, both random and zero-value faults resulted in linear performance degradation, with translation layers showing a greater resilience. For "Complete" injections, scale layers were robust against random noise but were the most vulnerable to zero-value faults at higher injection rates. Random injections strongly impacted scale layers in "Partial" configurations and translation layers in "Complete" configurations. Activation functions also played a critical role (Figure \ref{layer_outputs_activations_zeros_random}). Random injections severely affected translation layers, especially the final linear FC layers, while scale layers exhibited greater resilience. Zero-value injections caused exponential degradation in ReLU-activated scale layers, while translation layers remained steady across injection rates.

\subsection{\textbf{Bit-flips}}
\label{bitflipssection}

\begin{figure}[b] 
    \centering
    \includegraphics[width=1\columnwidth]{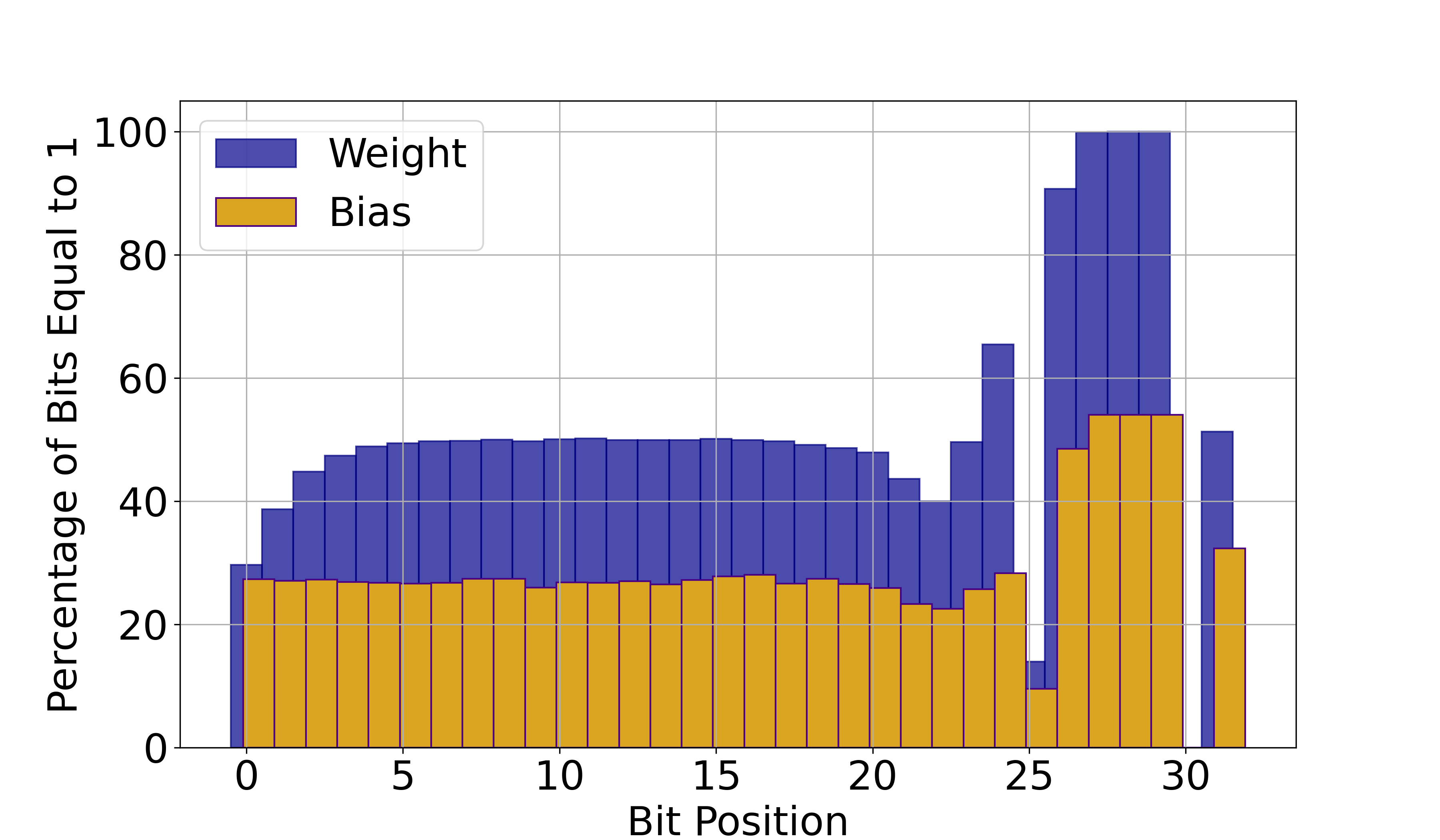}
    \captionsetup{justification=centering}
    \caption{Distribution of bits equal to 1 for each bit position in Physics-Informed Real NVP Model variables.}
    \label{bitsdistribution}
\end{figure}

\begin{figure}[t]
    \centering
    \begin{subfigure}[t]{0.24\textwidth}
        \centering
        \includegraphics[width=1\textwidth]{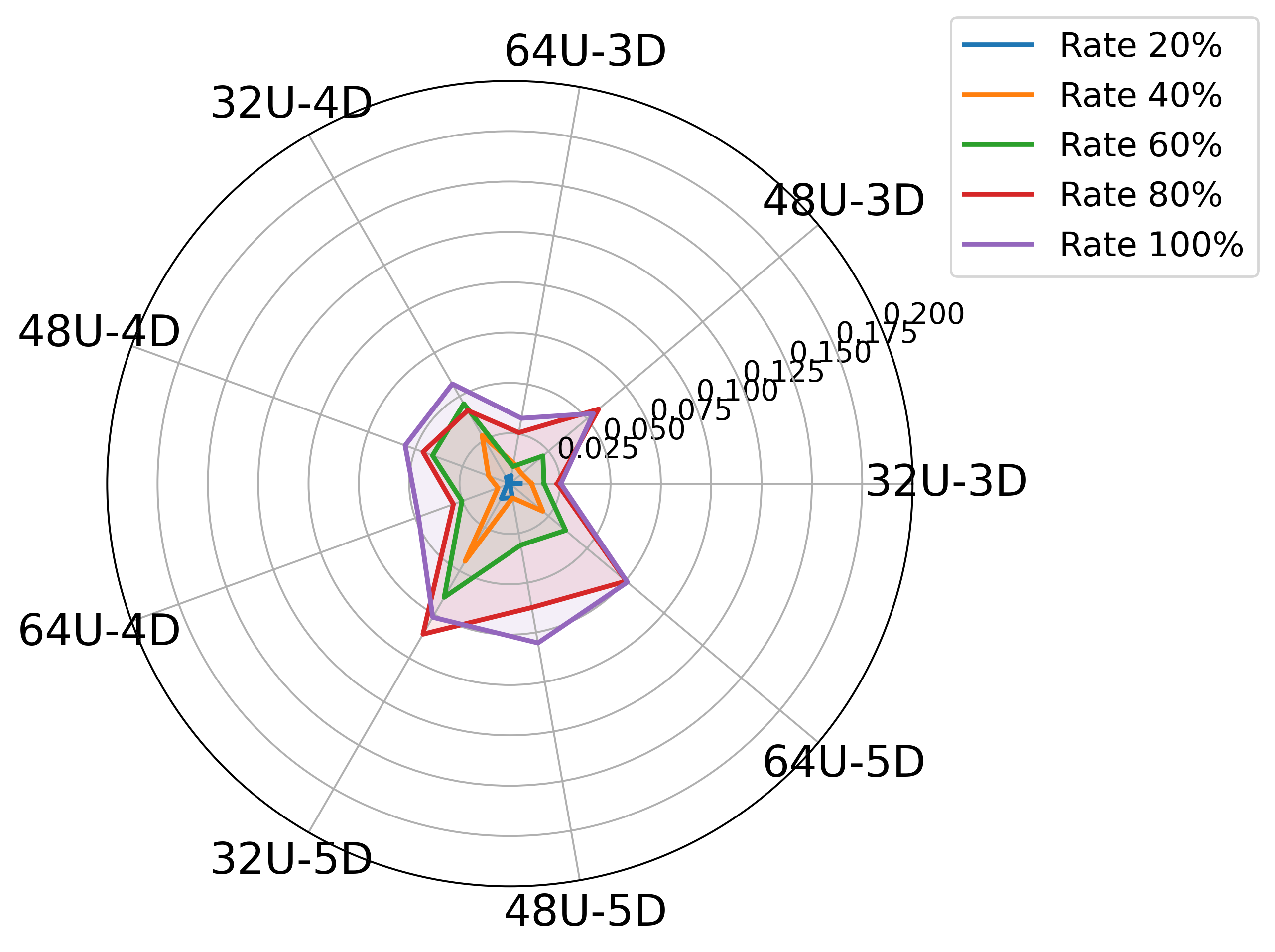}
        \captionsetup{font=footnotesize, justification=centering}
        \caption{Zeros injections with 4 coupling layers.}
        \label{stateszeros4models}
    \end{subfigure}
    \begin{subfigure}[t]{0.24\textwidth}
        \centering
        \includegraphics[width=1\textwidth]{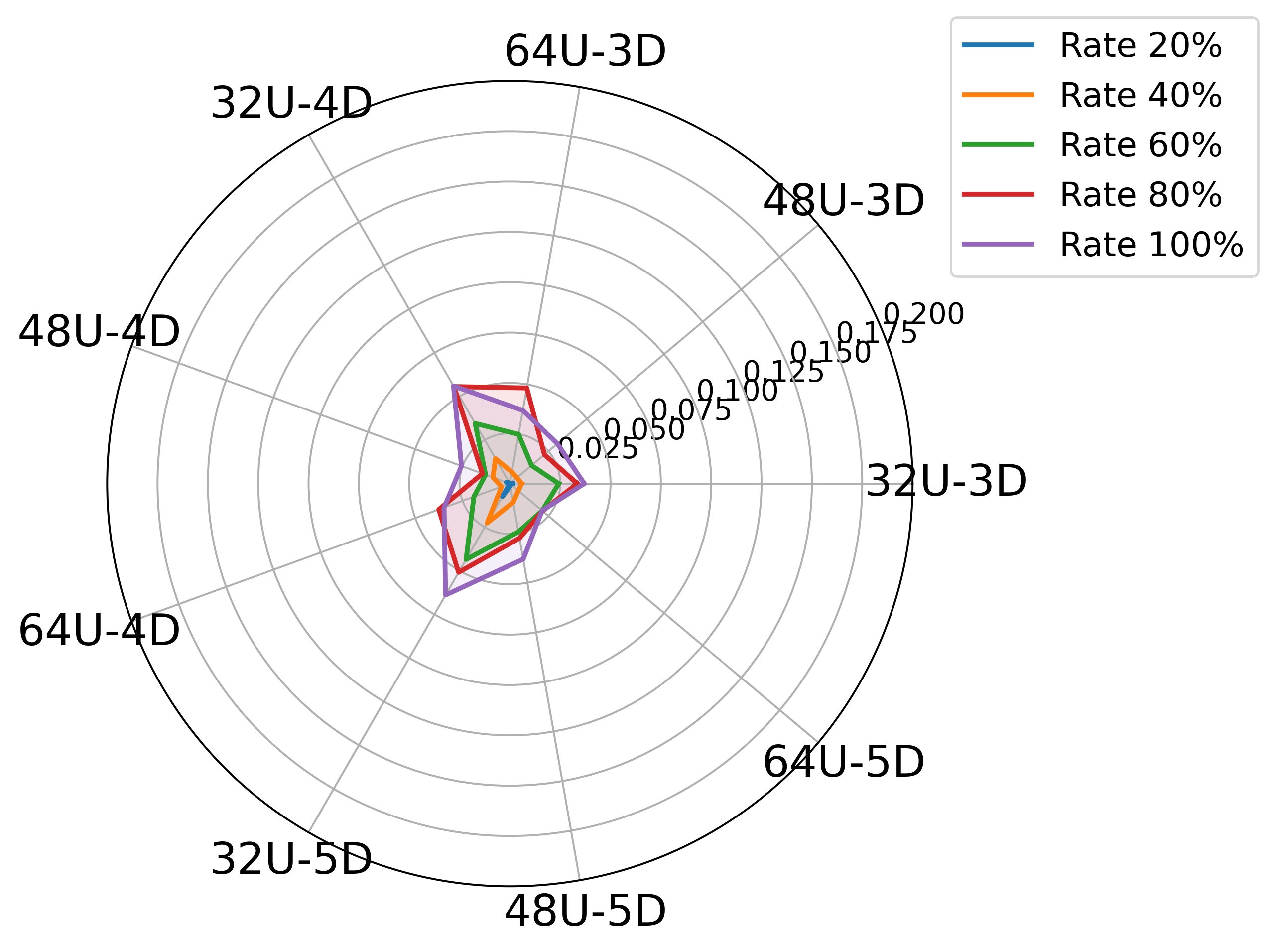}
       \captionsetup{font=footnotesize, justification=centering}
        \caption{Zeros injections with 6 coupling layers.}
        \label{stateszeros6models}
    \end{subfigure}
    \begin{subfigure}[t]{0.24\textwidth}
        \centering
        \includegraphics[width=1\textwidth]{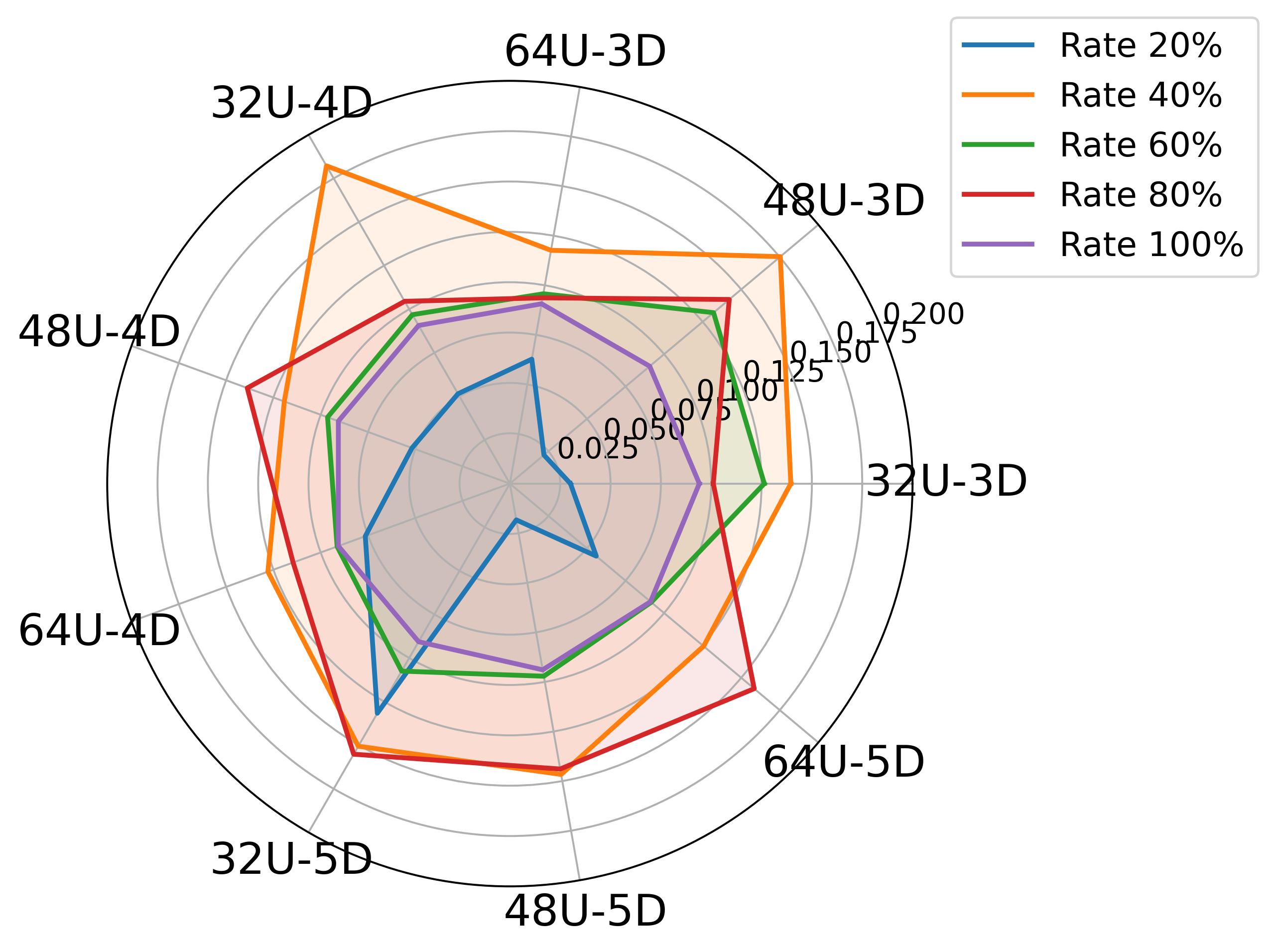}
        \captionsetup{font=footnotesize, justification=centering}
        \caption{Random injections with 4 coupling layers.}
        \label{statesrandom4models}
    \end{subfigure}
    \begin{subfigure}[t]{0.24\textwidth}
        \centering
        \includegraphics[width=1\textwidth]{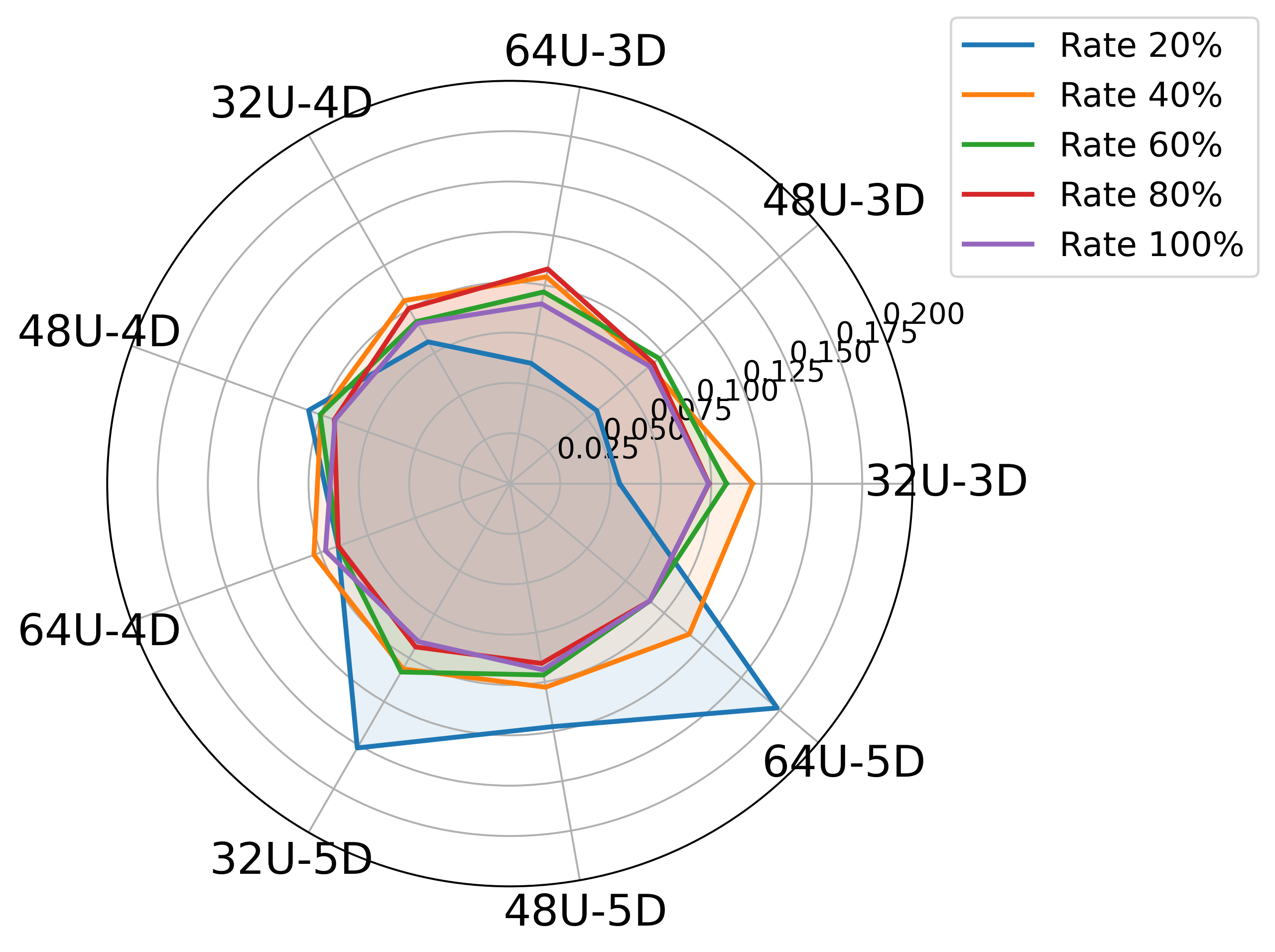}
        \captionsetup{font=footnotesize, justification=centering}
        \caption{Random injections with 6 coupling layers.}
        \label{statesrandom6models}
    \end{subfigure}
    \caption{Layer States injection across 18 Models with a fixed injection rate of 10\% and different variable layer percentage.}
    \label{state_models}
\end{figure}

\textbf{Layer States:} In Figure \ref{statesbitflips}, we observe the injection of bit-flips in all directions and in both positive and negative variables. Performance degradation begins at the 20th bit and follows a Gaussian distribution, peaking at the 25th bit in the exponent region, reflecting the critical role of specific bit positions in the Real NVP numerical stability. This finding aligns with the distribution of variables shown in Figure \ref{bitsdistribution}, emphasizing the importance of understanding bit-level behavior in designing fault-tolerant systems, especially for safety-critical applications. Flips in the mantissa bits show no performance degradation, while flips in bias have minimal impact, indicating the dominant influence of weights. However, the exponent bits (24–30) present key vulnerabilities due to their distribution. For instance, bit 30 is always set to 0, and flipping it to 1 can cause catastrophic transitions, turning values into infinity or NaN, leading to severe computational errors. This behavior is critical, as experimental data reveal that most fatal faults arise from such flips. Finally, an analysis of the weights showed that they have a slightly higher proportion of negative values, making the network marginally more sensitive to flips affecting these values. Flips from 1 to 0 generally cause minimal degradation and no system failures, whereas flips from 0 to 1 result in slightly higher performance degradation. These findings underscore the necessity of considering bit-level precision to mitigate the risks posed by such errors.

\begin{figure}[t]
    \centering
    \begin{subfigure}[t]{0.24\textwidth}
        \centering
        \includegraphics[width=1\textwidth]{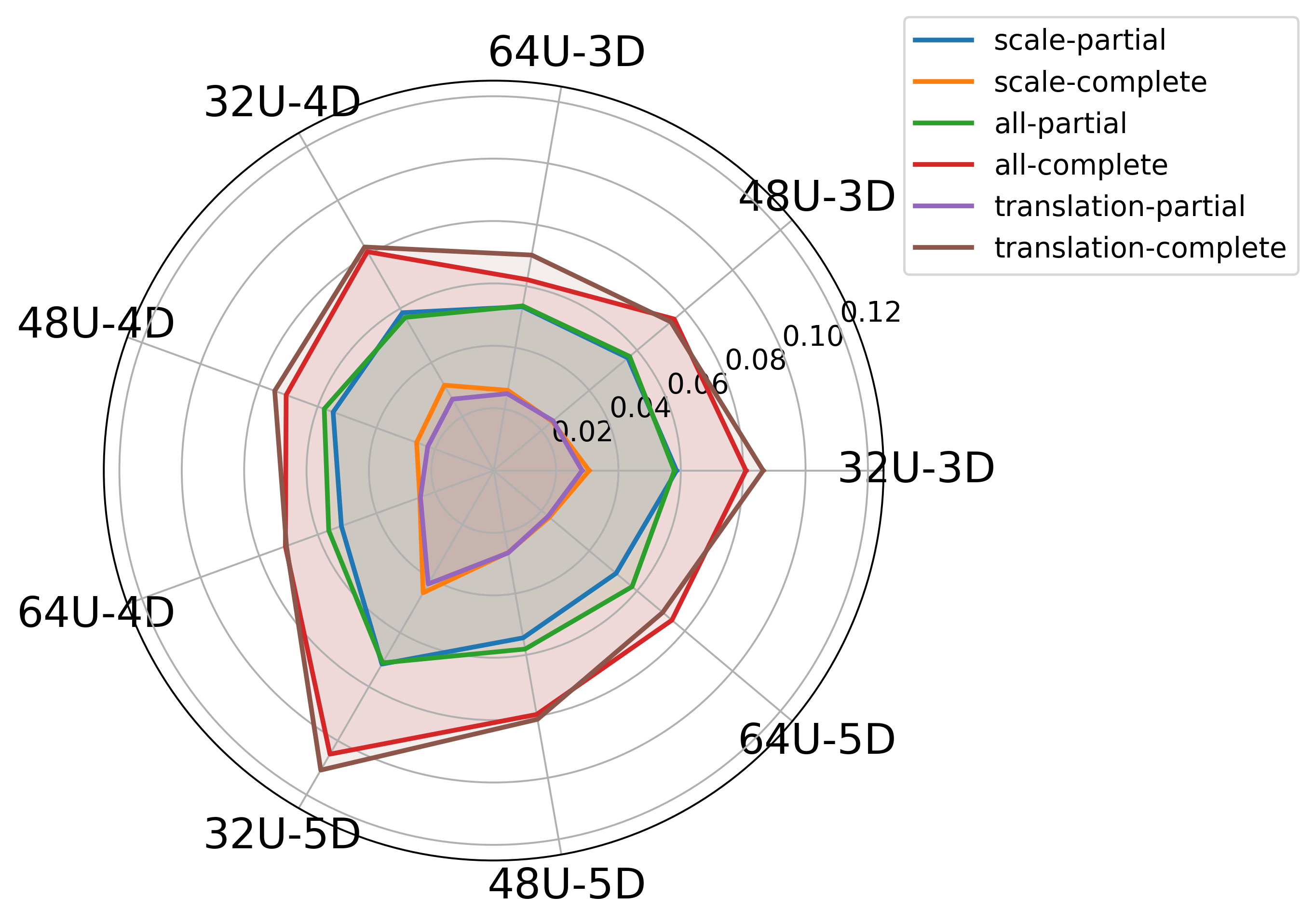}
        \captionsetup{font=footnotesize, justification=centering}
        \caption{Zeros injections with 4 coupling layers.}
        \label{outputszeros4models}
    \end{subfigure}
    \begin{subfigure}[t]{0.24\textwidth}
        \centering
        \includegraphics[width=1\textwidth]{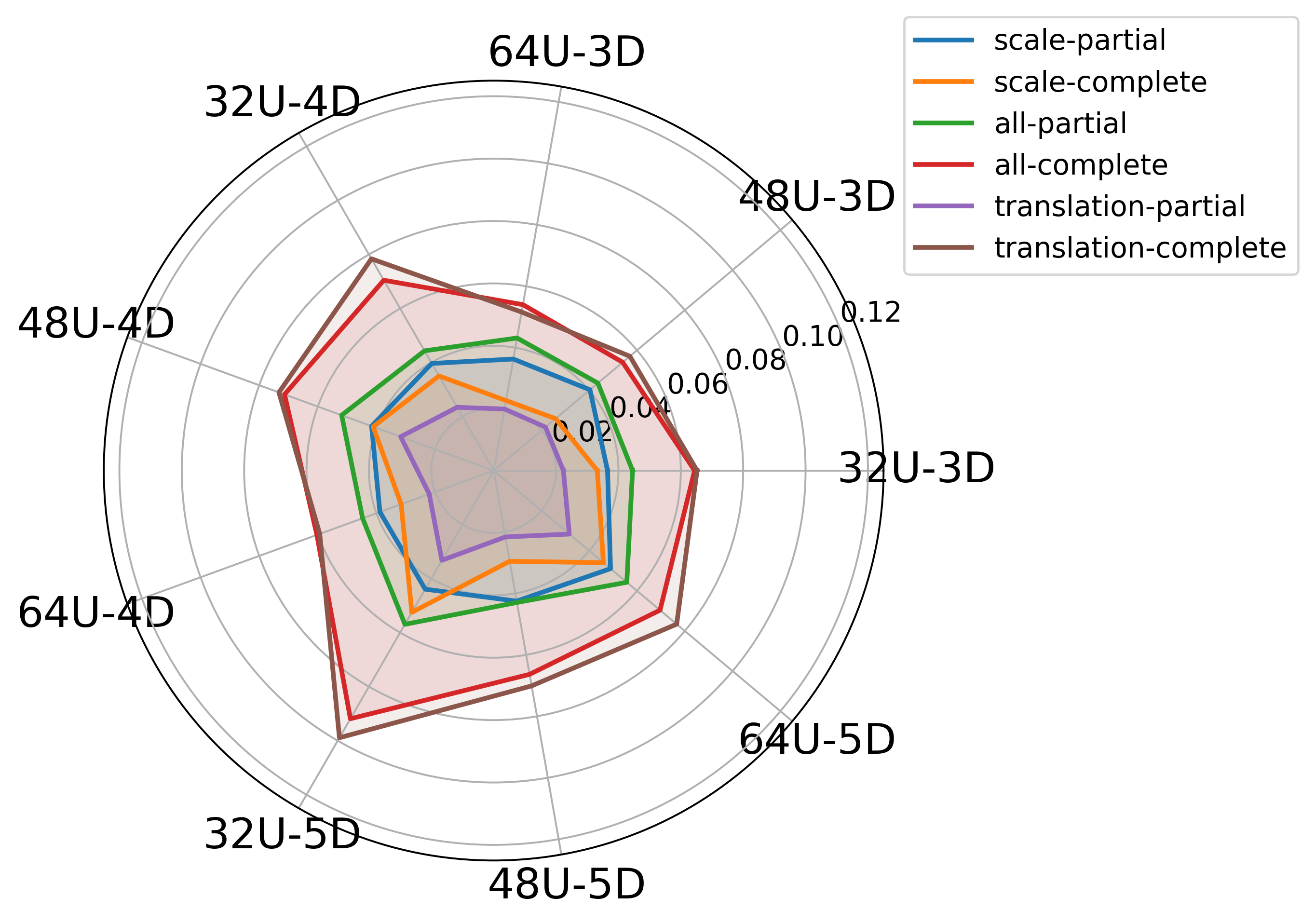}
        \captionsetup{font=footnotesize, justification=centering}
        \caption{Zeros injections with 6 coupling layers.}
        \label{outputszeros6models}
    \end{subfigure}
    \begin{subfigure}[t]{0.24\textwidth}
        \centering
        \includegraphics[width=1\textwidth]{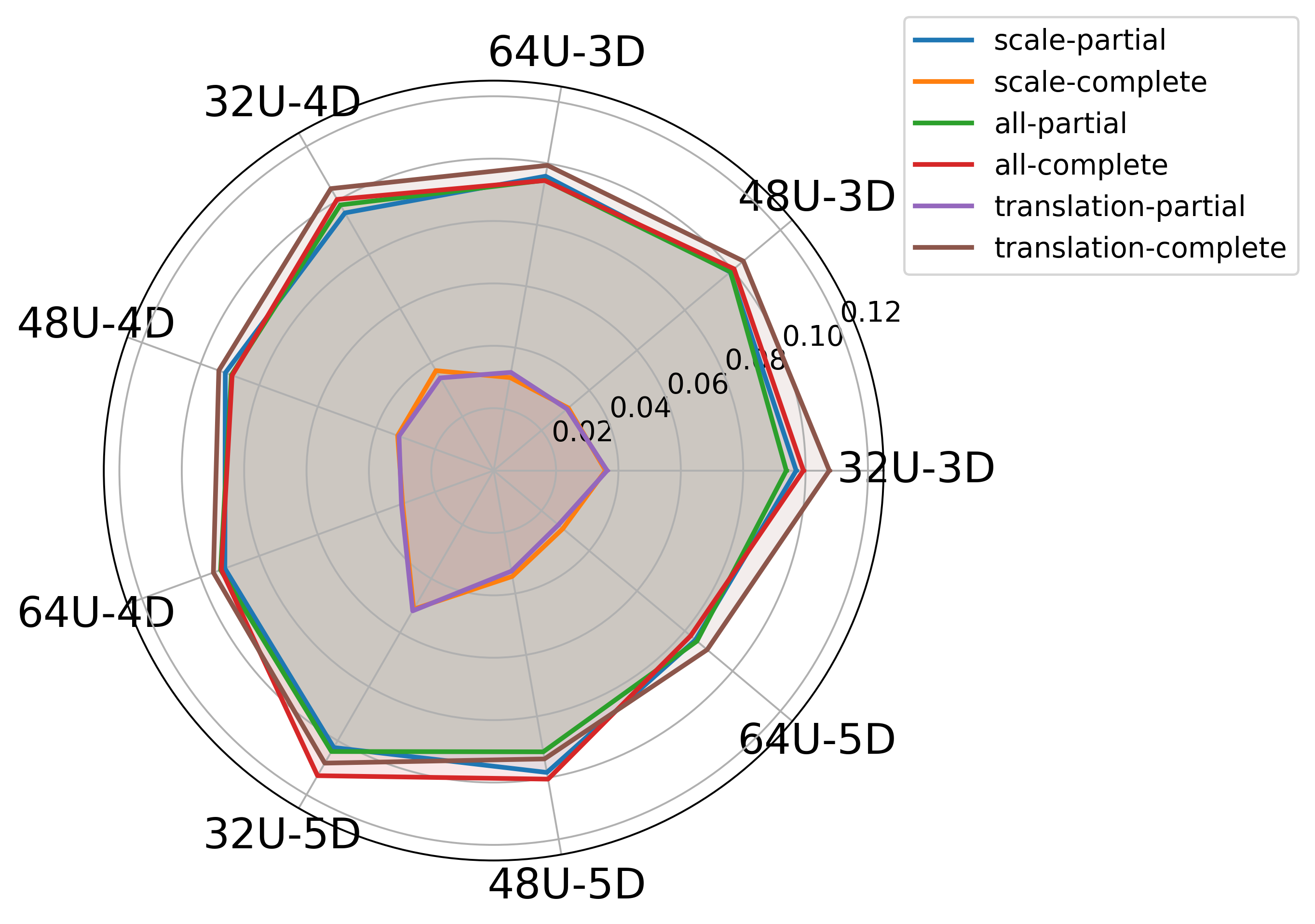}
        \captionsetup{font=footnotesize, justification=centering}
        \caption{Random injections with 4 coupling layers.}
        \label{outputsrandom4models}
    \end{subfigure}
    \begin{subfigure}[t]{0.24\textwidth}
        \centering
        \includegraphics[width=1\textwidth]{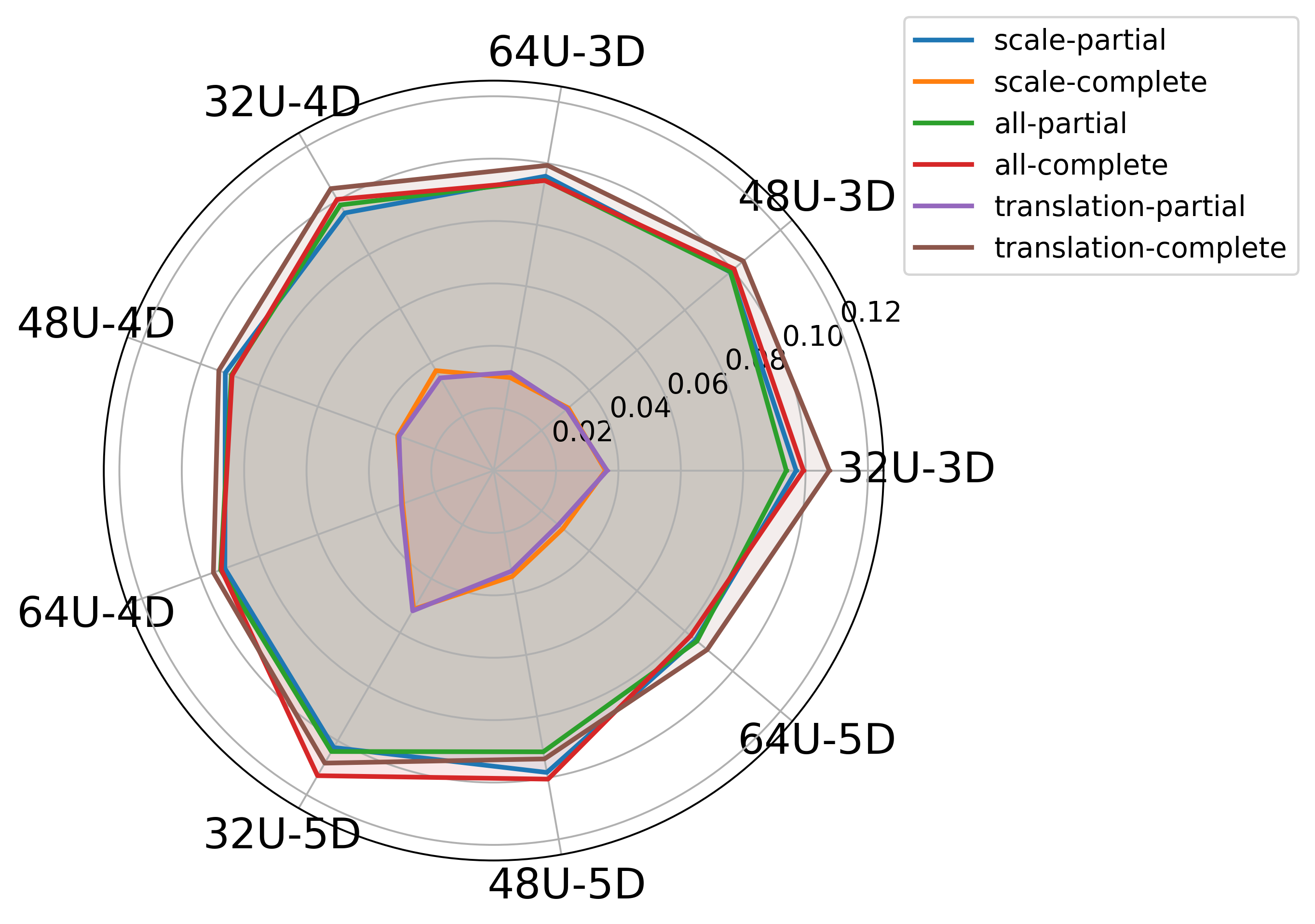}
        \captionsetup{font=footnotesize, justification=centering}
        \caption{Random injections with 6 coupling layers.s}
        \label{outputsrandom6models}
    \end{subfigure}
    \caption{Layer Outputs injections into different layers variables across 18 models with a fixed injection rate of 10\%.}
    \label{outputs_models}
\end{figure}

\textbf{Layer Output:} Figures \ref{outputsbitflipscomplete} and \ref{outputsbitflipspartial} examined the effects of bit-flips on scale and translation layers, both individually and in combination ('all'). For "Partial" injections, the model's performance was closely aligned with the behavior of scale layers, while "Complete" injections exhibited behavior associated with translation layers. Consistent with prior findings from Zeros and Random injections, translation layers showed resilience when injections occurred in the final layer after linear activation, whereas scale layers were more robust when "Complete" injections targeted hidden layers. Figure \ref{bitflipsactivations} shows the failure rates across various activation functions, finding that the scale tanh activation function was the most robust. These results emphasize the impact of model size, depth, and activation functions on bitflip-induced failures, providing insights into the Real NVP vulnerabilities and the effectiveness of different configurations in mitigating such errors.

\subsection{\textbf{Multiple Models}}
\label{multiplemodels}

\begin{figure}[t] 
    \centering
    \includegraphics[width=\columnwidth]{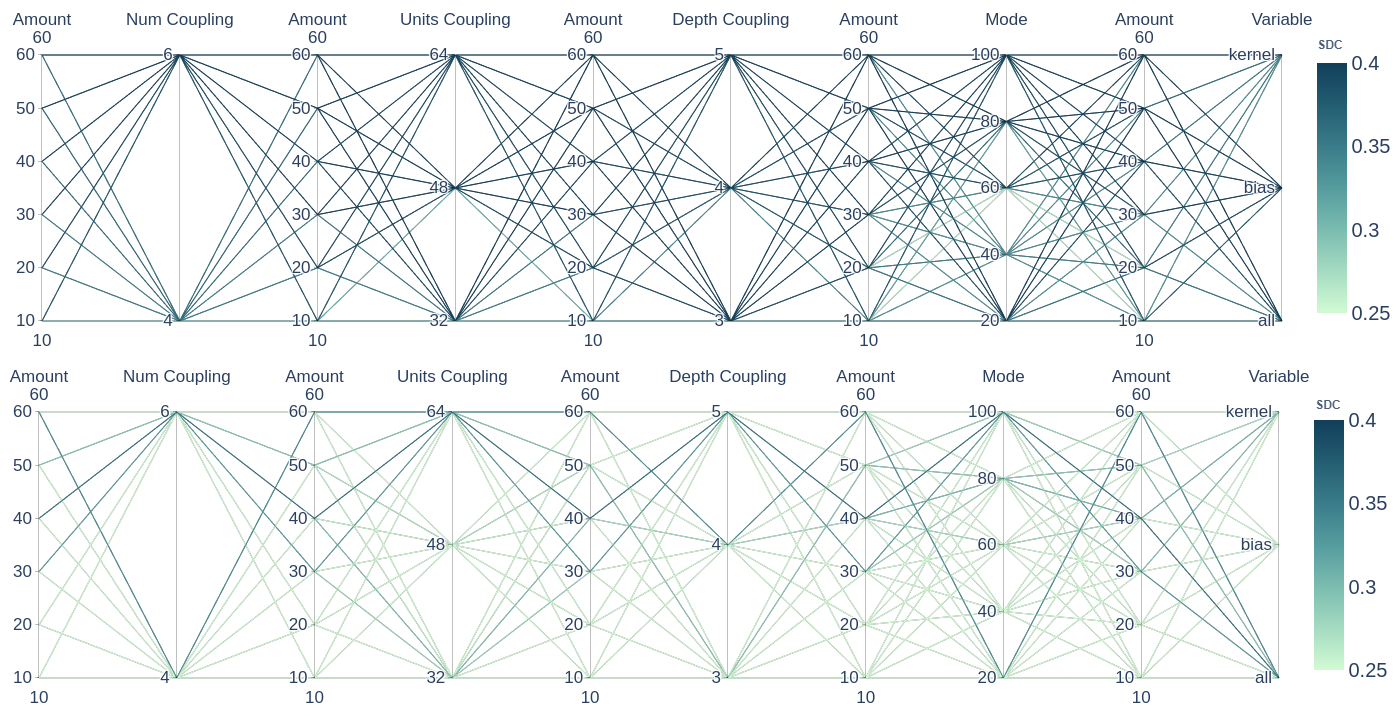}
    \captionsetup{justification=centering}
    \caption{Random (up) and Zeros (down) layer states experiments with all the configurations tested.}
    \label{layer_states_parallel}
\end{figure}

\begin{figure}[t] 
    \centering
    \includegraphics[width=\columnwidth]{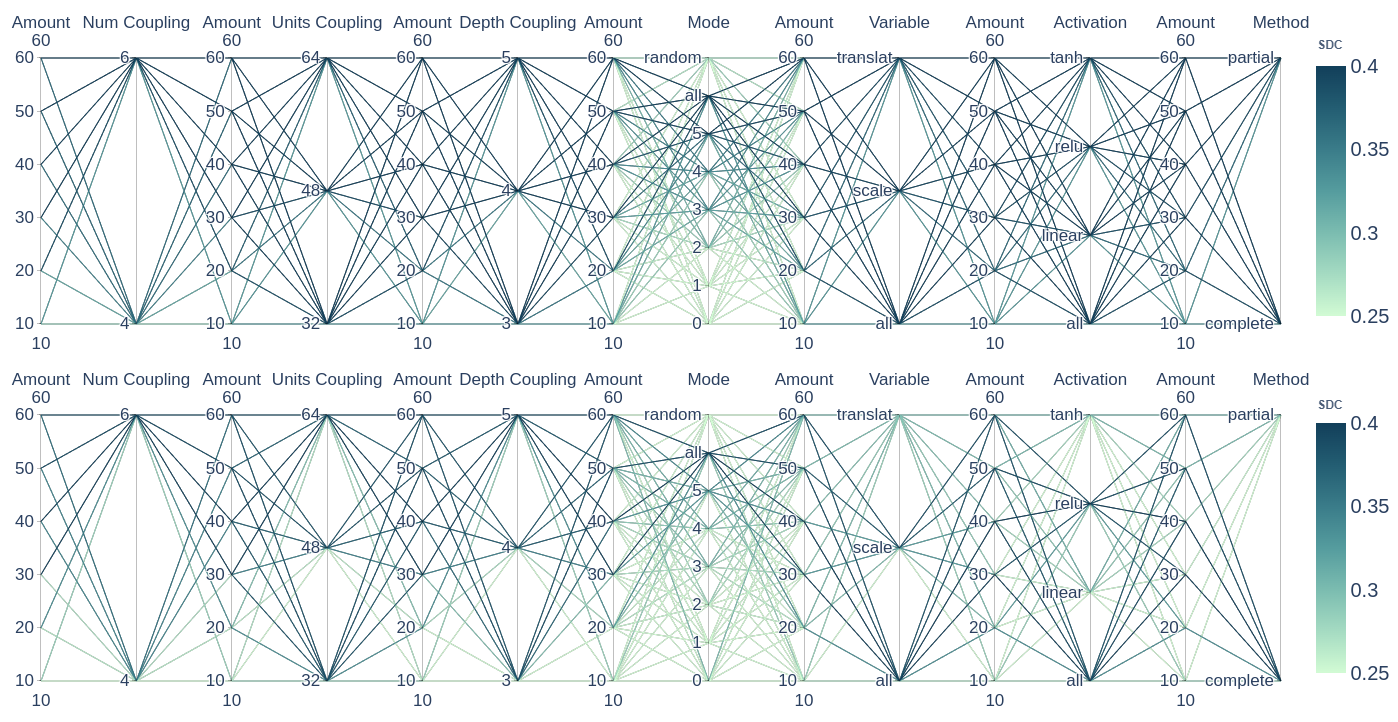}
    \captionsetup{justification=centering}
    \caption{Random (up) and Zeros (down) layer outputs experiments with all the configurations tested.}
    \label{layer_outpus_parallel}
\end{figure}

\begin{figure}[t] 
    \centering
    \includegraphics[width=\columnwidth]{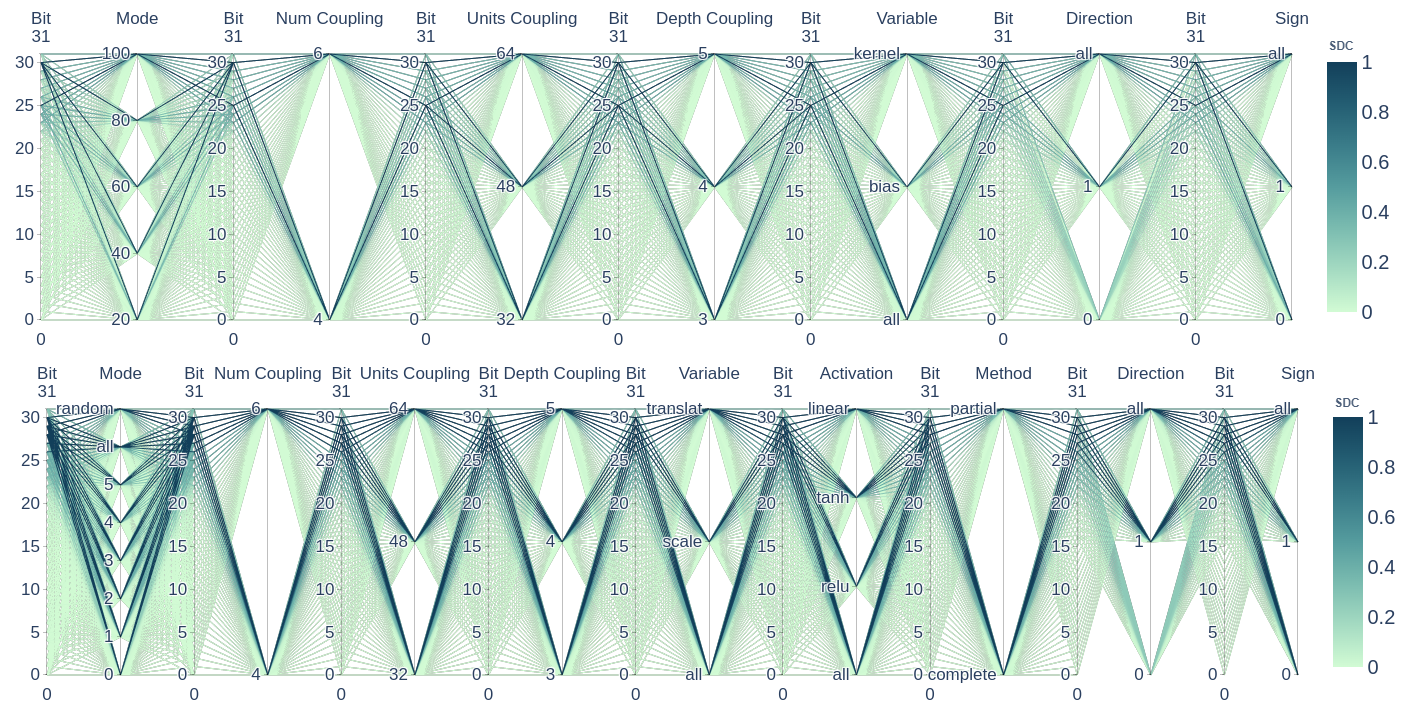}
    \captionsetup{justification=centering}
    \caption{Bit-flips experiments for layer states (up) and outputs (down) with all the configurations tested.}
    \label{bitflips_parallel}
\end{figure}

This analysis extends the investigation of injection effects to the 18 models referenced in section \ref{ls_setup}. Radial charts (Figures \ref{state_models} and \ref{outputs_models}) were employed to compare model performance across different injection scenarios. In these plots we use "D" and "U" to represent respectively the number of FC layers in each coupling layer and the number of units of each FC layer. Circular plots indicate consistent performance among the models, whereas deviations highlight variability in response. Figures \ref{layer_states_parallel}, \ref{layer_outpus_parallel}, and \ref{bitflips_parallel} present the complete results of the fault injection experiments conducted on all models in parallel coordinates plots. In these figures, darker lines represent higher SDC rates, indicating worse performance. The names of the axes refer to the parameters introduced in \ref{sec:fault_inj}.

\textbf{Layer States:} Figures \ref{state_models} and \ref{layer_states_parallel} show that Random injections lead to higher SDC rates with respect to Zeros. The changes in layer states using the Random approach show similar behaviors across all the experiments, leading to a higher SDC rate. In contrast, the Zeros approach yields a lower SDC rate compared to random. Notable results are observed when examining the "Variable" axis, which corresponds to changes in the weights, biases, or both. Similar behaviors are shown using the Random approach, whereas small variations are observed for the Zeros approach. Weight changes produce a higher SDC rate compared to bias changes, and the highest SDC rate is observed when both variables are altered simultaneously. For zero-value injections, models with deeper layers and fewer coupling units exhibited poorer performance, especially as the layer injection percentage increased. However, increasing the number of coupling units mitigated this degradation (Figures \ref{stateszeros4models} and \ref{stateszeros6models}). Bias injections had minimal impact on performance except in deep models with fewer units, while weights were more susceptible to performance loss, particularly in models with a higher number of units. When both bias and weights were injected, small changes in performance were observed for deep models with a lower amount of units. This highlights the importance of balancing model depth and the number of units to ensure robust performance under faults. For random-value injections similar trends were observed. Models with 4 coupling layers displayed chaotic performance at certain injection rates, such as 40\%, indicating sensitivity to small perturbations. Conversely, models with 6 coupling layers showed stable performance except at 40\%, where deeper configurations performed worse. Bias injections caused negligible effects across models, except for those with a large number of units.

\textbf{Layer Outputs:} Similar behaviors are observed when comparing the Random and Zeros approaches, Figures \ref{outputs_models} and \ref{layer_outpus_parallel}, but distinct differences emerge depending on injection type and model configuration. For zero-value injections, lower SDC rate values are achieved with small injections. On the "Mode" axis, higher SDC rate values are observed when injections target the last layers of the NN, with the highest SDC rate recorded when all output layers are injected. On the "Activation" axis, randomized outputs produce consistent behaviors across different activation functions. In contrast, the Zeros approach results in lower SDC rate values for "tanh" and "linear" activations compared to "relu." From Figures \ref{outputszeros4models} and \ref{outputszeros6models}, models with fewer units perform worse under Zeros injection, particularly those located at the edges of the graphs. Smaller hidden layers result in information loss, making it harder for models to capture relevant features. Deeper coupling layers exacerbate this issue as errors propagate more extensively through the network. However, models with larger numbers of units mitigate these effects, retaining more information and reducing error propagation. For random-value injections (Figures \ref{outputsrandom4models} and \ref{outputsrandom6models}), performance patterns closely resemble those observed with zero-value injections but exhibit key differences. The noise introduced by random values reduces performance, as reflected by the more rounded shapes in the graphs. Despite this reduction, the separation between models remains limited, with a notable outlier being the model featuring 32 units and a depth of 5. Random noise disrupts translation layers, but its effects are mitigated by scale layers, particularly in the combined "all" configuration.

\textbf{Bit-flips:} As shown in Figure \ref{bitflips_parallel}, higher exponent bits consistently result in the highest SDC rate across both configurations. For layer states bit 30 and the peak of the Gaussian at bit 25 exhibit a strong correlation with SDC rate (Figure \ref{statesbitflips}). Alterations in these positions lead to higher SDC rate values. In contrast, bit 31, associated with the sign, has a comparatively smaller impact. Layer outputs exhibit similar behavior to layer states but demonstrate poorer performance across all exponent bits, ultimately contributing to higher SDC rate values. This behavior was found in all experiments.

\section{Conclusions \& Future Work}
This research developed a fault injection framework tailored for Real-NVP networks, designed for space applications where robustness, resilience, and compactness are critical. Our customized framework, inspired by existing solutions, like TensorFI \cite{tensorfi}, supports fault injections in two main areas: layer states, targeting network variables before inference, and layer outputs, enabling real-time injections during inference. These configurations allow for both broad testing and granular ablation studies, helping to identify the network's strengths and vulnerabilities. The study explored three methods of fault injection: Zeros, which sets variables or outputs to zero; Random, simulating random fluctuations; and Bit-flips, mimicking bit-level alterations caused by radiation. By analyzing faults in both the variables and the outputs, we evaluated the network's ability to classify anomalies in multivariate time-series under space-like conditions. The results offer valuable insights into improving the robustness of Real NVP networks for safety-critical applications. Future work should focus on expanding fault injection frameworks that adopt a hybrid approach, combining software/hardware-level injection methods. This will provide a more comprehensive understanding of the reliability of AI systems in challenging environments like space.

\section*{Acknowledgment}
This work has been developed with the contribution of the Politecnico di Torino Interdepartmental Centre for Service Robotics (PIC4SeR https://pic4ser.polito.it).

\printbibliography

\vspace{12pt}

\end{document}